\PassOptionsToPackage{table}{xcolor}
\documentclass[10pt,letterpaper]{fairmeta}

\usepackage[T1]{fontenc}
\usepackage{lmodern}
\usepackage{microtype}
\usepackage{amsmath}
\usepackage{graphicx}
\usepackage{caption}
\usepackage{algorithm}
\usepackage[noend]{algpseudocode}
\usepackage{listings}
\usepackage{booktabs}
\usepackage{tabularx}
\usepackage{longtable}
\usepackage{float}
\usepackage[table]{xcolor}
\usepackage{tikz}
\usetikzlibrary{arrows.meta}
\usepackage{enumitem}
\usepackage{needspace}
\usepackage{xurl}
\usepackage[numbers,sort&compress]{natbib}
\usepackage[colorlinks=true,linkcolor=blue!45!black,citecolor=blue!45!black,urlcolor=blue!55!black]{hyperref}

\hypersetup{
  pdftitle={RoboScope Kit: Unified Interfaces for Fast and Scalable Embodied-AI Evaluation},
  pdfauthor={Deng-Ke Wu; Yifan Jin; Yida Yang; Lei Qian; Zhaoye Zhou; Hongru Zhang; Kai Ye; Shuyuan Hu; Yiran Mao; Zewei Ye; Bo Zhao},
  pdfsubject={Technical report on embodied-AI evaluation infrastructure}
}

\newcommand{\system}{RoboScope Kit}
\newif\ifshowrev\showrevfalse
\newcommand{\rev}[1]{\ifshowrev{\leavevmode\color{blue}#1}\else#1\fi}
\newcommand{\revcolor}{\ifshowrev\color{blue}\fi}
\newcommand{\code}[1]{\texttt{#1}}
\definecolor{evoblue}{HTML}{145F91}
\definecolor{evolight}{HTML}{EAF2F8}
\newcolumntype{Y}{>{\raggedright\arraybackslash}X}
\newcolumntype{P}[1]{>{\raggedright\arraybackslash}p{#1}}
\graphicspath{{figures/}}
\algrenewcommand\algorithmicrequire{\textbf{Input:}}
\algrenewcommand\algorithmicensure{\textbf{Output:}}
\title{RoboScope Kit: Unified Interfaces for Fast and Scalable Embodied-AI Evaluation}
\author[1]{Deng-Ke Wu}
\author[1,2]{Yifan Jin}
\author[1,3]{Yida Yang}
\author[1,4]{Lei Qian}
\author[1,5]{Zhaoye Zhou}
\author[1]{Hongru Zhang}
\author[1]{Kai Ye}
\author[1]{Shuyuan Hu}
\author[1,6]{Yiran Mao}
\author[1]{Zewei Ye}
\author[1]{Bo Zhao}
\affiliation[1]{School of AI, Shanghai Jiao Tong University}
\affiliation[2]{Central South University}
\affiliation[3]{Tongji University}
\affiliation[4]{ShanghaiTech University}
\affiliation[5]{Chinese Academy of Agricultural Sciences}
\affiliation[6]{Nanjing University}
\date{\rev{November 21, 2026}}
\correspondence{Bo Zhao at \email{bo.zhao@sjtu.edu.cn}}
\metadata[Project]{\url{https://mint-sjtu.github.io/RoboScope-Kit/}}
\metadata[Code]{\url{https://github.com/MINT-SJTU/RoboScope-Kit}}

\abstract{
\rev{Evaluating embodied-AI policies in closed loop across simulation benchmarks such as LIBERO and RoboTwin requires
repeated integration, as each benchmark defines its own evaluator, observation layout, and action convention.
Evaluation can also be slow: with a StarVLA checkpoint, the 5,000 episodes of RoboTwin Easy take 17.5 hours on four
RTX PRO 5000 GPUs. Profiling the official StarVLA evaluators identifies two bottlenecks: alternating between one
simulator and inference leaves the GPU at only 14.3\% average utilization on LIBERO-Long, while rendering at every
control step on RoboTwin means an action-chunk policy reads only one of every 16 rendered observations.} We
present \textbf{\system{}}, an evaluation toolkit \rev{that integrates each model and benchmark once through a unified
interface, keeping each benchmark's native evaluator and scoring, and accelerates evaluation through two mechanisms.
\textbf{(1) Scheduling:} simulator workers and policy replicas scale independently, with autotuning sizing both
pools to each workload and host.
\textbf{(2) Runtime optimization:} the simulation runtime skips unconsumed simulation and rendering work without
changing the policy's inputs or outcomes.
On a single RTX 4090,} \system{} \rev{runs} LIBERO-Long \textbf{\boldmath 6.2$\times$} faster and 500 RoboTwin Easy
episodes \textbf{\boldmath 13.2$\times$} faster \rev{than StarVLA, and 2.1--7.3$\times$ faster than RLinf and
vla-eval. Scaling to four RTX PRO 5000 GPUs, it evaluates all 5,000 RoboTwin Easy episodes in 66 minutes
(\textbf{\boldmath 15.8$\times$} faster). LIBERO-Long success rates stay within 0.3 percentage
points of StarVLA, and every RoboTwin episode matches its exact outcome and terminal step.}
}

\begin{document}

\maketitle

\begingroup
\setlength{\intextsep}{6pt}
\begin{figure}[t]
\centering
\includegraphics[width=\linewidth]{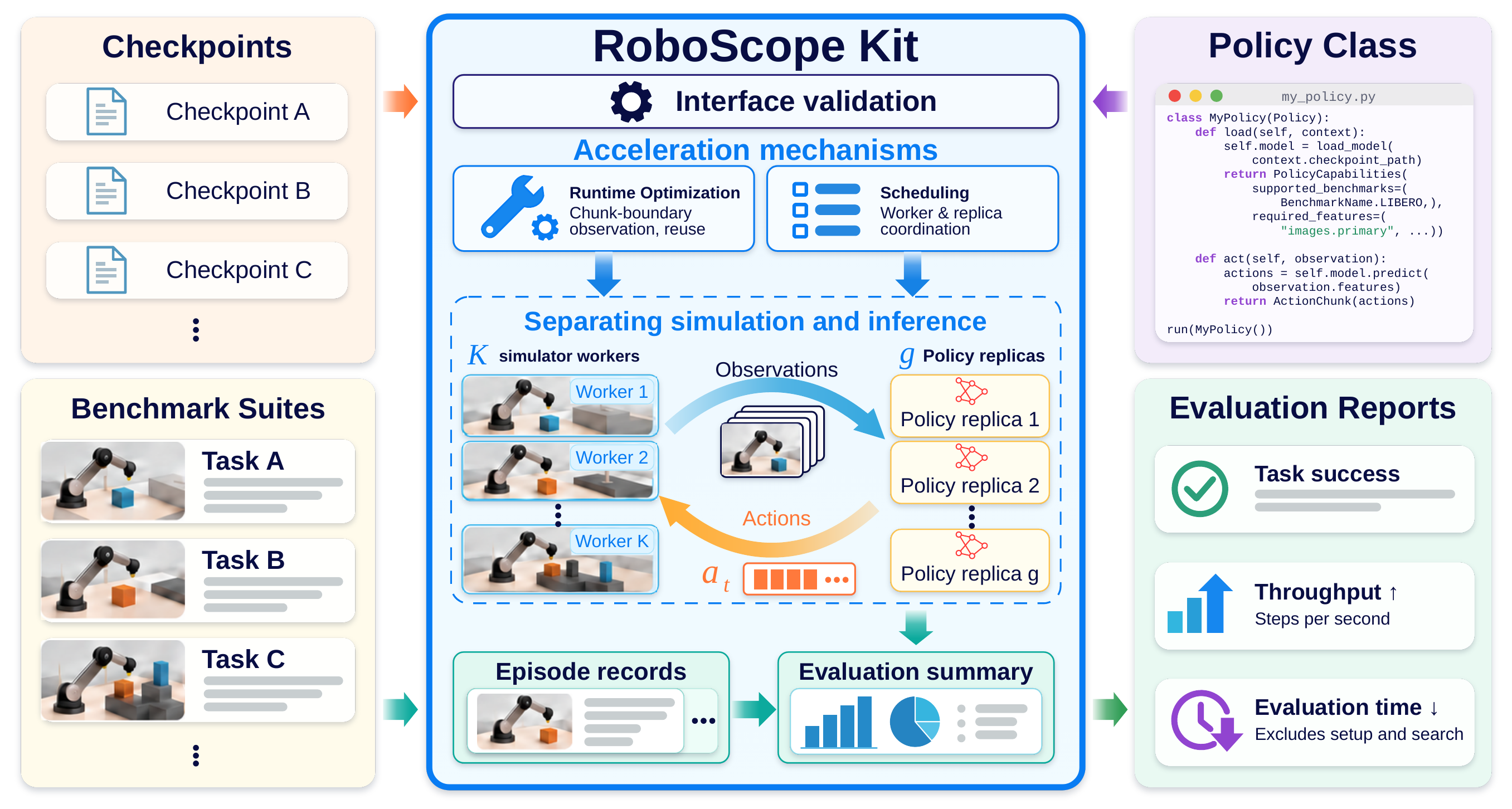}
\caption{\system{} connects Policies and benchmark adapters. After interface validation, acceleration
has two parts: runtime optimization removes unconsumed simulation work inside each worker, and scheduling
coordinates $K$ simulator workers that exchange observations and actions with $g$ Policy replicas.
Each episode yields one episode record. The Policy class holds the model-side processing.}
\label{fig:v2-overview}
\end{figure}
\endgroup

\section{Introduction}
\label{sec:v2-intro}
Simulation benchmarks evaluate complementary aspects of robot behavior: knowledge transfer in
LIBERO \citep{libero2023}, bimanual manipulation in RoboTwin \citep{robotwin22025},
language-conditioned task sequences in CALVIN \citep{calvin2022}, and household manipulation in
RoboCasa \citep{robocasa2024}. Each benchmark brings its own evaluator, observation layout, action
convention, and episode bookkeeping, and each policy family brings its own preprocessing and
inference loop. Connecting several policies to several benchmarks repeats this integration for every
pair. Execution cost compounds the problem: with a StarVLA checkpoint, the official evaluation of
500 LIBERO-Long episodes takes about 50 minutes on one RTX 4090, \rev{and the official evaluation of the 5,000
episodes of RoboTwin Easy takes 17.5 hours on four RTX PRO 5000 GPUs.} Comparing checkpoints, seeds, or training stages multiplies these costs, and a shared evaluation
layer can remove much of this repeated work.

Profiling the StarVLA evaluators shows where this time goes (Section~\ref{sec:v2-time-profile}).
Simulation occupies CPU cores while one loaded model leaves the GPU mostly idle, so simulators and
models are best scaled separately. On RoboTwin, an action-chunk Policy reads one of every 16 rendered
observations, so the others need not be rendered.

\rev{We propose \system{}, an evaluation toolkit that removes repeated integration through a unified interface
and accelerates evaluation through two mechanisms, while keeping each benchmark's own rules and
scoring. The unified interface integrates each model and each benchmark once. Scheduling scales simulators and
loaded models independently to keep both CPU cores and GPUs busy, and runtime optimization removes redundant
simulation work without changing any episode's outcome.}

This report makes \rev{two} contributions:
\begin{enumerate}[itemsep=0.2em,topsep=0.3em]
  \item \textbf{A shared evaluation interface.} A model is wrapped once as a Policy class and
  then runs on any benchmark that provides the inputs it needs, without changes to \system{}.
  With a StarVLA checkpoint, \system{} with one simulator worker
  and one Policy replica reproduces all 500 LIBERO-Long success outcomes of the official StarVLA evaluation.
  \item \textbf{Fast evaluation through scheduling and runtime optimization.} \rev{Scheduling runs simulator
  workers and Policy replicas as separate pools, so several simulators share each loaded model, and a time-bounded
  autotuner sizes both pools for each workload and host. Runtime optimization removes simulation work that
  neither feeds the Policy nor affects the score. On one RTX 4090, \system{} evaluates LIBERO-Long
  \textbf{\boldmath 6.2$\times$} faster and 500 RoboTwin Easy episodes \textbf{\boldmath 13.2$\times$} faster than the
  official StarVLA evaluation, and 2.1--7.3$\times$ faster than vla-eval and RLinf. On four RTX PRO 5000 GPUs,
  \system{} evaluates the 5,000 RoboTwin Easy episodes \textbf{\boldmath 15.8$\times$} faster. Success rates on LIBERO-Long are
  equivalent, and every RoboTwin episode has the same outcome and terminal step.}
\end{enumerate}

\section{Background and Motivation}
\label{sec:v2-motivation}
\subsection{Simulation-Based Policy Evaluation}
Simulation benchmarks let researchers compare robot policies on fixed task distributions at a cost
and scale that physical trials cannot reach. An evaluation runs a fixed set of episodes. Each episode
starts from a benchmark-defined task and initial state, alternates between rendering an observation,
querying the policy, and stepping the simulator with the returned action, and ends when the task
succeeds or a step limit is reached. The benchmark then scores the episode records, usually as the
fraction of successful episodes.

Benchmarks differ at every stage of this path. \rev{LIBERO supplies two $256\times256$ cameras, 16-D proprioception, and a language instruction, and expects a 7-D end-effector delta with a gripper command. RoboTwin supplies three cameras and dual-arm joint state and expects a 14-D absolute joint and gripper target. CALVIN and RoboCasa use further layouts (Appendix~\ref{app:v2-contracts}).}
Termination rules, step limits, and scoring differ as well. Two policies, or two evaluation systems,
are comparable when they run the same episodes under the same rules, so for every run in this
report we record the episode list, checkpoint, software versions, and hardware.

\subsection{Where Evaluation Time Goes}
\label{sec:v2-time-profile}
\begin{figure}[tbp]
\centering
\includegraphics[width=\linewidth]{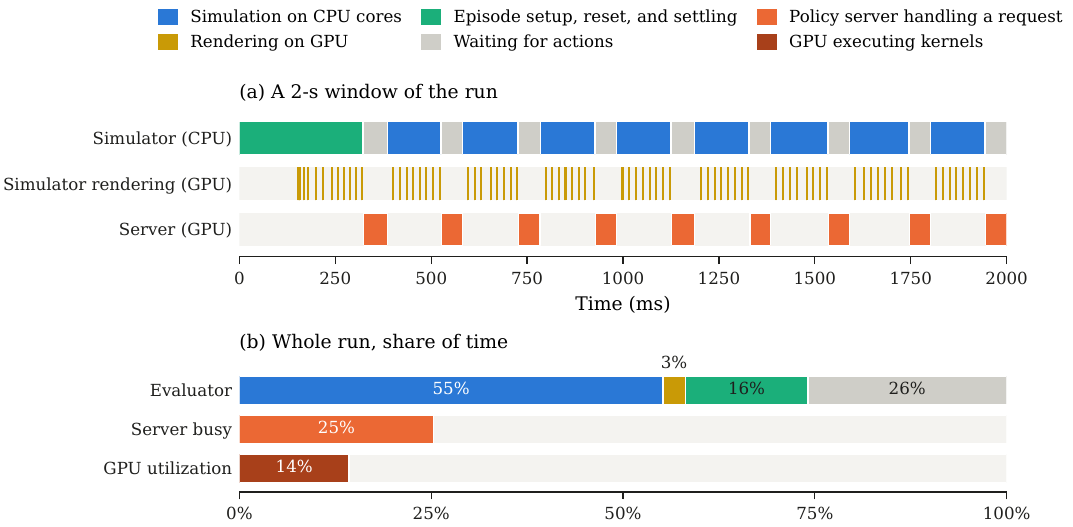}
\caption{The StarVLA LIBERO evaluator and its policy server on 50 LIBERO-Long episodes
(five per task) on one RTX 4090. (a) Execution timeline over a 2-s window. \rev{The simulator runs in the evaluator process, with physics and control on CPU cores and camera rendering on the GPU (each tick is one render of about 0.4\,ms). The policy server also runs on the GPU.} (b) Over the whole run: evaluator time,
the share of time the policy server is handling a request, and mean GPU utilization (the share of
time a GPU kernel is executing).}
\label{fig:v2-native-timeline}
\end{figure}

\paragraph{Simulation, rendering, and inference costs.}
An episode's wall time divides among simulator initialization and reset, physics stepping,
rendering, observation transfer, Policy inference, and cleanup. These stages use different
resources: physics, robot control, and environment logic use CPU cores, rendering uses the GPU's
graphics pipeline, and inference uses GPU compute and memory. Figure~\ref{fig:v2-native-timeline}
traces the StarVLA evaluator and its policy server on 50 LIBERO-Long episodes on one
RTX 4090.

The evaluator runs one simulator and queries the server once per action chunk, so simulation and
inference take turns. {\revcolor
The evaluator's time divides as follows (Figure~\ref{fig:v2-native-timeline}b):
\begin{itemize}[itemsep=0.1em,topsep=0.3em]
  \item \textbf{Simulation on CPU cores} takes 55\%, which covers simulation steps and the per-step handling of
  observations and actions. Most of it is robot control and MuJoCo forward kinematics, computed at each of the 25
  physics substeps per control step, and physics integration takes 11\%.
  \item \textbf{Rendering} the two cameras on the GPU takes 3\%.
  \item \textbf{Episode setup, reset, and settling} take 16\%.
  \item \textbf{Waiting for actions} takes 26\%. An inference takes 55\,ms, of which the GPU executes kernels for
  26\,ms, because the model issues about 3,100 kernel launches per inference and the GPU waits between them.
\end{itemize}
}
\noindent The server is busy 25\% of the time, and the GPU averages 14.3\% utilization
(Appendix~\ref{app:v2-time-profile}). Neither the CPU cores nor the GPU is kept busy. Since simulators consume CPU cores
and Policy replicas consume GPU memory and compute, the two can be scaled separately.

\needspace{9\baselineskip}
{\revcolor
The StarVLA RoboTwin evaluator, profiled in the same way on one RTX PRO 5000 over 100 RoboTwin Easy
episodes, all ten of every fifth task, spends its time differently (Table~\ref{tab:v2-robotwin-profile}):
\begin{itemize}[itemsep=0.1em,topsep=0.3em]
  \item \textbf{Rendering} takes 43.5\% of the evaluator's time, against 3\% on LIBERO. The evaluator renders \rev{four}
  cameras at every control step, \rev{of which the Policy reads three,} and a render takes about 42\,ms.
  \item \textbf{Control and physics} take 23.6\% and 12.9\%.
  \item \textbf{Waiting for actions} takes 4.2\%, because one inference supplies 16 control steps.
  \item \textbf{Per-episode and per-task overhead} takes the rest: \rev{the expert check of each seed} takes
  6.9\% and scene setup 5.2\%, and restarting the policy server for each task adds 5.3\% of the wall time.
\end{itemize}
}

\paragraph{Unconsumed observations in action-chunk policies.}
Many current policies predict action chunks: one inference returns $H$ actions, which the evaluator
executes over the next $H$ control steps before querying the Policy again. The StarVLA RoboTwin
checkpoint has $H=16$. The StarVLA RoboTwin evaluator still renders all \rev{four} cameras at every
control step, so 15 of every 16 rendered observations are never read\rev{: in the profile above, 2,346 of
37,064 rendered observations reach the Policy, and the unread 93.7\% account for most of the 43.5\% of
evaluator time spent rendering}. Rendering only at chunk
boundaries keeps physics stepping, action execution, and success checks at the native control
frequency, and the Policy receives exactly the observations it reads.

\subsection{Design Goals}
These observations lead to five design goals.
\textbf{(G1) Unified.} Integrate each model and each benchmark once, and reuse it with every counterpart.
\textbf{(G2) Faithful.} Reproduce the outcomes of each benchmark's official evaluation, under its own
tasks, initial states, termination, and scoring.
\textbf{(G3) Efficient.} Keep both CPU cores and GPUs busy, and skip simulation work that changes
neither the Policy's inputs nor the score.
\textbf{(G4) Automatic.} Configure each evaluation for its workload and host without manual tuning,
and report what this costs.
\textbf{(G5) Scalable.} Extend one evaluation from one GPU to several GPUs and instances.

\section{Design}
\label{sec:v2-design}
\subsection{Overview}
\system{} has four parts (Figure~\ref{fig:v2-overview}). A \emph{Policy} is a class, installed in
its model's own environment, that loads a checkpoint and keeps the model's preprocessing. A
\emph{benchmark adapter} wraps the benchmark's \emph{native evaluator}, which keeps task definitions,
initial states, termination, and scoring (G2, faithful). \rev{Two runtimes, the \emph{Control Runtime} and the \emph{Simulation Runtime}, connect the Policy and the
benchmark adapter. The Control Runtime, which schedules the evaluation, expands} a benchmark
suite into a \emph{manifest}, the ordered list of \emph{episode keys} (task, episode, seed), and
groups the keys into \emph{work items}. It starts $K$ \emph{simulator workers}, processes that each
own one simulator, and $g$ \emph{Policy replicas}, processes that each hold one loaded Policy, and
collects one \emph{episode record} per episode key. \rev{The Simulation Runtime} runs inside each
simulator worker: the native evaluator, instrumented by the adapter, steps the simulator, renders
observations, and sends them to a Policy replica. \emph{Runtime optimization}
in the Simulation Runtime \rev{speeds up evaluation by removing} simulation work the Policy never consumes
(\S\ref{sec:v2-runtime}). \emph{Scheduling} in the Control Runtime \rev{also speeds it up by separating} simulation from
inference, \rev{choosing} $K$ and $g$, and \rev{sharding} work across instances (\S\ref{sec:v2-architecture},
\S\ref{sec:v2-selection}, \S\ref{sec:v2-sharding}).

An evaluation proceeds in four stages. The Control Runtime first validates that the Policy's declared
capabilities match the benchmark's contract. It then runs \emph{autotuning}, short probes that choose $K$ and $g$,
unless they are fixed. Next it runs the manifest as work items across $K$ workers that send observations to
$g$ replicas. Finally it aggregates completed items into benchmark scores and execution metrics.

\subsection{Evaluation Interface}
\label{sec:v2-policy}
\paragraph{Observation and action contracts.}
A model is integrated once as a Policy and a benchmark once as an adapter, and each is reused with
every counterpart (G1, unified). Each benchmark specification names its observation features and its action. \rev{A feature has a canonical name such as \code{images.head}, \code{state.joints}, or \code{language.instruction}, a kind, a dtype, and an optional shape. The action has a name, a dimension, and a dtype.} A Policy
declares the features it requires, the suites it supports, the chunk lengths it returns, its
action representation, and an observation stride per benchmark. Checkpoint-specific work, including
image resizing and cropping, normalization, tokenization, and the transform from model outputs to
benchmark actions, stays inside the Policy.
Adapters apply returned actions in the benchmark's native convention and execute action chunks one
step at a time at the native control frequency.

\paragraph{Policy lifecycle and episode state.}
A Policy replica moves through five operations, one of which also has a batched form (Table~\ref{tab:v2-lifecycle}). \rev{\code{load} reads one checkpoint and returns the Policy's capabilities, and \code{reset} allocates state for one episode. \code{act} maps an observation to an action or action chunk, and its batched form \code{act\_batch}, which defaults to calling \code{act} per observation, serves several at once. \code{end\_episode} releases that episode's state, and \code{unload} releases the model.} Several simulator workers share a replica, so
requests from different episodes interleave. Every request carries an episode identifier, and stateful
Policies key their histories, caches, and action queues by it. \rev{Model weights and other replica-scope resources live from load to unload, while episode-scope state lives from reset to end of episode (Figure~\ref{fig:v2-lifecycle}).}

\paragraph{Interface validation.}
Before any rollout, the Control Runtime checks that the requested benchmark and suite are among the Policy's
declared capabilities, that every required feature is provided, and that the chunk length and action
representation resolve to one contract across all replicas. During rollouts it checks each
observation's fields, dtypes, and fixed dimensions, and each returned action's numeric type, shape,
and dimension.

\subsection{Separating Simulation and Inference}
\label{sec:v2-architecture}
Each simulator worker is a process that owns one simulator instance and the benchmark's
dependencies. Each Policy replica is a process that loads one model on an assigned GPU in the model's own
environment, so the replicas of one instance can spread over all of its GPUs (G5, scalable). Workers and replicas never share an interpreter, so benchmark and model
environments can pin conflicting libraries. \rev{The two form independent pools of sizes $K$ and $g$, which can be sized to keep both CPU cores and GPUs busy (G3, efficient). The notation $K8/g2$ denotes eight workers sharing two replicas.} Workers send observations to replicas over local
request--response connections and receive action chunks, and every message carries the episode
identifier. Because workers share replicas, GPU memory holds $g$ copies of the model however many
simulators run. All processes of an evaluation belong to one process group, which the Control Runtime stops
as a unit at shutdown or on failure.

\subsection{Autotuning}
\label{sec:v2-selection}
\paragraph{Probe workloads and resource constraints.}
The autotuner chooses $K$ and $g$ automatically within a time budget (G4, automatic),
and Appendix~\ref{app:v2-search-time} reports its cost.
Concurrency is chosen by running the actual workload. A probe starts $K$ workers and $g$ replicas
with the real Policy, runs a short episode set, and measures throughput $q$ in control steps per
second, consumed CPU cores $c$, GPU utilization $u$, CPU throttling $t$, and per-replica GPU memory
$v$. Throughput is measured in steady state: each worker receives three work items, and the estimate
drops each worker's first item, which includes environment creation and Policy warmup, and the items
that finish after the first worker runs out of work. A probe fails when a worker is missing or fails,
an out-of-memory (OOM) error occurs, its episodes do not finish within the time limit, or its success rate
falls below 85\% of the first probe's on the episodes both probes ran.
Resources bound the candidates: CPU cores bound the worker count, and the GPU memory left after
loading replicas bounds the replica count.

\paragraph{Selection rules and search budgets.}
\rev{Algorithm~\ref{alg:v2-selection}, the \emph{diagonal search}, first adds simulator workers alone and then adds workers and replicas together, a diagonal in the $(K, g)$ plane. Table~\ref{tab:v2-selection-rules} defines the predicates it uses.} The \emph{worker phase} keeps one replica per GPU and doubles $K$ while a
doubling raises throughput by at least 25\%, a quarter of linear scaling. The first doubling that gains
less marks the \emph{knee} of the throughput curve, beyond which more workers add little. Past the knee,
the replicas limit throughput. The accepted $K$ per replica, $r$, is the number of workers one replica keeps busy.
The \emph{replica phase} then adds one replica per GPU at a time and scales the workers with it,
$K = r g$, keeping a step while it raises throughput by at least $\varepsilon$. Both phases stop before
CPU, GPU, or throttling limits are exceeded, and when a replica step's predicted GPU memory does not fit, the replica phase lowers $K$ in steps of
the GPU count, down to the current $K$, until it fits, and stops if none fits. Each probe is limited to $\tau_c$ and the whole search to $\tau$. A probe is \emph{accepted} when it
completes, stays within the CPU, GPU, and throttling limits, and gains enough over the current configuration:
25\% in the worker phase and $\varepsilon$ in the replica phase. The search then moves to the accepted
configuration and tries the next one. \rev{A probe that misses any of these conditions is \emph{rejected} and recorded with its reason. After a rejection, the search keeps the last accepted configuration and goes on to the replica phase or stops (Algorithm~\ref{alg:v2-selection}).} When the budget runs out, the last accepted configuration also stands.

\begin{algorithm}[tbp]
\caption{Diagonal search for autotuning.}
\label{alg:v2-selection}
\begin{algorithmic}[1]
\Require Workload $W$, Policy $P$, host with $G$ GPUs
\Ensure Configuration $(K^\ast, g^\ast)$, or $\bot$ if the first probe fails
\State $(K^\ast, g^\ast, m^\ast) \gets (G, G, \Call{Probe}{G, G})$
\If{$m^\ast$ fails} \Return $\bot$ \EndIf
\While{$K^\ast < \textsc{WorkerCap}(m^\ast)$} \Comment{worker phase}
  \State $K \gets \min(2K^\ast, \textsc{WorkerCap}(m^\ast))$;\ $m \gets \Call{Probe}{K, G}$
  \If{$m$ fails \textbf{or not} $\textsc{WithinLimits}(m)$} \Return $(K^\ast, g^\ast)$ \EndIf
  \If{\textbf{not} $\textsc{Scales}(m, m^\ast)$} \textbf{break} \EndIf
  \State $(K^\ast, m^\ast) \gets (K, m)$
\EndWhile
\State $r \gets K^\ast / G$
\While{$g^\ast + G \le \textsc{ReplicaCap}$} \Comment{replica phase}
  \State $g \gets g^\ast + G$;\ $K \gets \max(K^\ast, \min(\mathrm{round}(r g), \textsc{WorkerCap}(m^\ast)))$
  \While{$K - G \ge K^\ast$ \textbf{and not} $\textsc{FitsMemory}(K, g)$} $K \gets K - G$ \EndWhile
  \If{\textbf{not} $\textsc{FitsMemory}(K, g)$} \textbf{break} \EndIf
  \State $m \gets \Call{Probe}{K, g}$
  \If{$m$ fails \textbf{or not} $(\textsc{WithinLimits}(m) \land \textsc{Gains}(m, m^\ast))$} \textbf{break} \EndIf
  \State $(K^\ast, g^\ast, m^\ast) \gets (K, g, m)$
\EndWhile
\State \Return $(K^\ast, g^\ast)$
\end{algorithmic}
\end{algorithm}

\begin{table}[tbp]
\centering
\small
\caption{Predicates and parameters of Algorithm~\ref{alg:v2-selection}. \rev{$C$ is the CPU quota in cores and $V$ the memory per GPU. The measurements of probe $m$ are $q(m)$, $c(m)$, $u(m)$, $t(m)$, and $v(m)$, and $K(m)$ is its worker count.}}
\label{tab:v2-selection-rules}
\renewcommand{\arraystretch}{1.25}
\begin{tabularx}{\linewidth}{@{}P{30mm} Y P{37mm}@{}}
\toprule
Name & Definition & Setting \\
\midrule
$\textsc{Probe}(K, g)$ fails &
A worker is missing or fails, an OOM error occurs, episodes do not finish within
$\min(\tau_c, \text{remaining } \tau)$, or success on the episodes shared with the first probe falls below $(1-\delta)$ times the first probe's &
$\tau_c = 15$\,min, $\tau = 30$\,min, $\delta = 0.15$ \\
$\textsc{WorkerCap}(m^\ast)$ &
$\max(G, \lfloor (C-1)/2 \rfloor)$ until two probes at $g = G$ have run, and $K^\ast + \lfloor (C - 1 - c(m^\ast))/\hat c \rfloor$ afterwards, capped at 64 in both cases, where $\hat c$ is the CPU cost per worker fitted from the probes at $g = G$ with the fewest and most workers &
-- \\
$\textsc{ReplicaCap}$ &
$\min(16, C, G\lfloor (V - r_v)/v \rfloor)$, with $v$ measured by the first probe &
$r_v = 2$\,GiB reserved per GPU \\
$\textsc{FitsMemory}(K, g)$ &
The busiest GPU's peak memory, extrapolated from the incumbent with the replica footprint and the fitted
memory per worker, stays below $V - r_v$ & -- \\
$\textsc{WithinLimits}(m)$ &
$c(m) \le C - 1$, $u(m) \le u_{\max}$, and $t(m) \le t_{\max}$ when $t$ is measured &
$u_{\max} = 0.95$, $t_{\max} = 0.40$ \\
$\textsc{Scales}(m, m^\ast)$ &
$q(m) \ge \bigl(1 + \kappa\,(K(m)/K^\ast - 1)\bigr)\, q(m^\ast)$ &
$\kappa = 0.25$ \\
$\textsc{Gains}(m, m^\ast)$ &
$q(m) \ge (1 + \varepsilon)\, q(m^\ast)$ &
$\varepsilon = 0.03$ \\
\bottomrule
\end{tabularx}
\end{table}

\subsection{Sharding Across Instances}
\label{sec:v2-sharding}
One host bounds the workers and replicas an evaluation can use. \rev{\system{} adds a second scheduling level above each instance's local $K/g$ Control Runtime (Figure~\ref{fig:v2-multi-instance}), so that one evaluation can span several instances (G5, scalable).} Given the
ordered work items $\mathcal{W}=(w_0,\ldots,w_{M-1})$, shard $i$ of $N$ runs
\[
  \mathcal{W}_i=\{w_j\in\mathcal{W}: j\bmod N=i\},\qquad 0\le i<N.
\]
The shards are disjoint and together form the requested workload, and each work item keeps its
canonical episode indices and initial states or goals. Each
instance runs its own queue, simulator workers, and Policy replicas and, when enabled, its own
autotuner.

Each shard writes its own result, progress counter, and build identity, and two merge paths combine
them. Record-tree merging checks that the episode records cover every work item exactly once and
then applies the benchmark's aggregation to the combined records. \rev{Result-only reduction checks that all shards share the benchmark protocol and build identity and then combines per-shard counts and partial aggregates. Official metrics that are not additive are recomputed from these counts.}

\subsection{Runtime Optimization}
\label{sec:v2-runtime}
\paragraph{Simulation Runtime reuse and redundant work removal.}
Runtime optimization removes only work that neither feeds the Policy nor affects the score
(G3, efficient), and chunk-boundary observation applies only at the stride the Policy declares.
Within a rollout, the RoboTwin Simulation Runtime removes work that repeats across episodes or only
validates seeds. It reuses one SAPIEN Engine and Renderer for all episodes a worker runs
instead of rebuilding them per episode. It also streamlines the control loop, which repeats identical computations
within each physics step. RoboTwin's native evaluator validates each seed with an expert check.
Given a verified seed cache, the runtime skips this check and, for joint-position actions, cuRobo
initialization. End-effector actions retain one cuRobo planner per worker. Chunk-boundary observation follows the
Policy's declared stride.

\paragraph{Chunk-boundary observation.}
A Policy declares an observation stride $s$ for a benchmark. \rev{At control step $t$, the simulator worker renders a new observation if the Policy runs inference at $t$, which happens every $H$ steps for chunk length $H$, or if $t$ is a multiple of $s$. Otherwise, the worker passes the previous observation.} With
$s=H$, rendering happens once per chunk and every inference receives the observation of its own step.
The default stride is 1, which reproduces the official cadence of one render per step. Randomized
lighting, the interactive viewer, video logging, and point-cloud observations keep one render per step
for any $s$. The StarVLA RoboTwin Policy declares $s=H=16$. The stride-16 run on four RTX 4090 GPUs matches the
stride-1 runs on one RTX 4090 on every success outcome and terminal step of the 500 RoboTwin Easy episodes
(Section~\ref{sec:v2-reference}).

\subsection{Work Distribution and Failure Handling}
\label{sec:v2-integrity}
\begin{figure}[tbp]
\centering
\includegraphics[width=\linewidth]{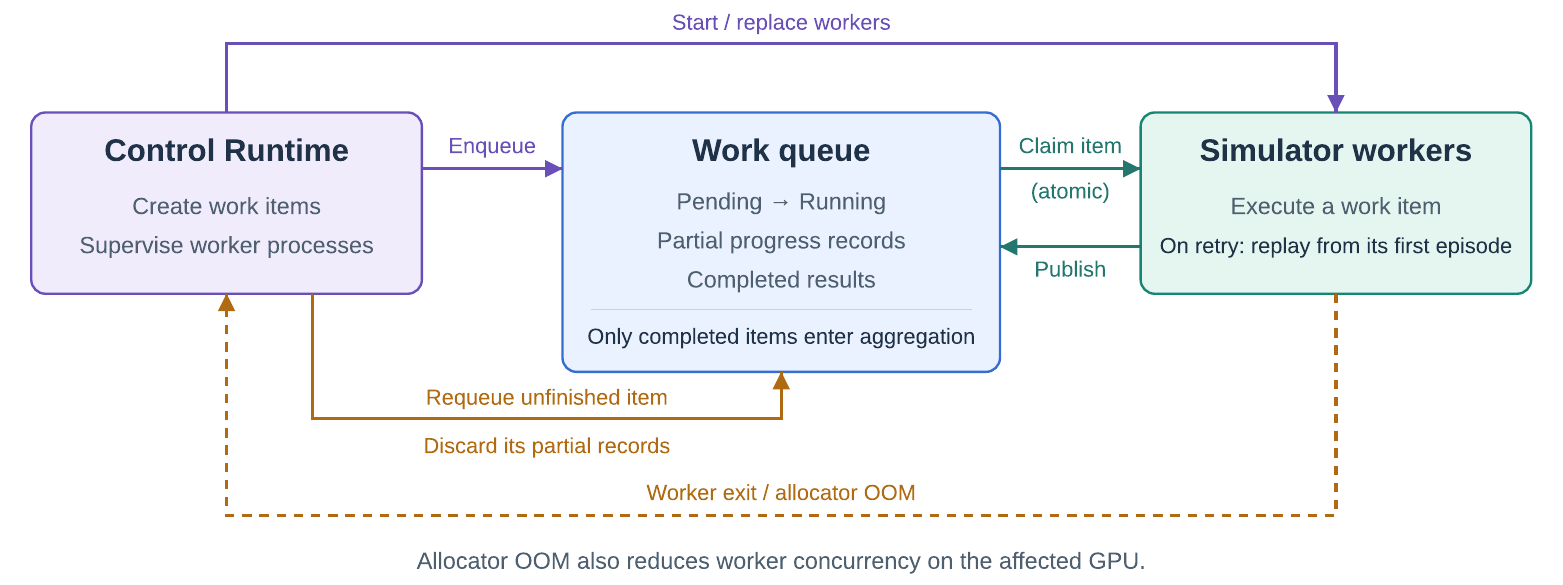}
\caption{Work distribution and worker recovery. Workers atomically claim queued items and publish
partial progress or completed results. After a worker exit or allocator OOM error, the
Control Runtime requeues unfinished work, discards its partial records, and starts a replacement worker,
which replays the item from its first episode. \rev{Only completed items enter aggregation, so a replayed episode is counted once.} Policy-replica
exit has a different failure boundary: it aborts the evaluation (Section~\ref{sec:v2-recovery}).}
\label{fig:v2-architecture}
\end{figure}

\paragraph{Work items and result integrity.}
The Control Runtime converts the ordered manifest into work items: one task's episodes for LIBERO, and consecutive
per-task episode windows for RoboTwin, interleaved across tasks so that long rollouts spread over
workers. Each item keeps the canonical episode indices and the seeds or initial states of its
episodes. A worker claims an item atomically, moving it from pending to running, appends terminal
records to the item's partial state as episodes finish, and publishes a completed result when the
item ends. Publication is idempotent: an identical repeated result is accepted, and a conflicting
result for the same item is an integrity error. \rev{Before scoring the completed items,} the
Control Runtime checks coverage, exactly one record per requested episode key with no duplicates or unexpected keys,
and that each record's seed matches the manifest. Figure~\ref{fig:v2-architecture} shows the item
states and the recovery path.

\paragraph{Failure handling.}
\label{sec:v2-recovery}
Worker failures are recovered at work-item granularity (Figure~\ref{fig:v2-architecture}). When a simulator worker exits or raises an
allocator OOM error, the Control Runtime returns its claimed item to pending, discards the item's
partial records, and starts a replacement worker, which replays the item from its first episode.
After an OOM error it also lowers worker concurrency on the affected GPU while the other GPUs
continue. Aggregation reads completed items, so a replayed episode contributes one record. \rev{A Policy replica that exits aborts the evaluation, after which the Control Runtime stops every process of the evaluation and checks that none remains.}

\section{Implementation}
\label{sec:v2-implementation}
\paragraph{Packaging and process layout.}
\system{} is distributed as a Python package that defines the Policy API and contains the Control Runtime,
the Simulation Runtime, and the benchmark adapters. A Policy is installed in its model's environment, and a configuration
file names the command that starts its replica process. The Control Runtime is installed at a pinned version
next to the Policy, with a checksummed benchmark configuration. Benchmark simulators, datasets, and
native evaluators come from shared storage when an evaluation starts. \rev{The command-line entry point takes a benchmark, suite, checkpoint, and output directory, runs autotuning and the formal evaluation, and writes the manifest, probe history, episode records, metrics, and resource samples to the output directory.} The Control Runtime, simulator workers, and Policy replicas run as separate processes on one host and communicate over
local Unix-domain sockets. For a sharded evaluation, the entry point also takes a shard index and
shard count. Our evaluation platform, a cluster service outside \system{}, runs each evaluation on up to four
instances: it launches and polls the instances, collects the shard reports, and publishes the score
once every shard has completed. If a shard fails, the evaluation fails and all of its instances are
released. Cross-instance sharding comes with a merge tool that checks that the shards are disjoint
and cover the workload and then recomputes the official result.

\begin{table}[t]
\centering
\small
\caption{\rev{Benchmarks and suites in the adapter catalog.}}
\label{tab:v2-compatibility}
\renewcommand{\arraystretch}{1.2}
\begin{tabular}{@{}l c l@{}}
\toprule
Benchmark & Suites & Suite names \\
\midrule
LIBERO & 5 & \code{libero\_spatial}, \code{libero\_object}, \code{libero\_goal}, \code{libero\_10}, \code{libero\_90} \\
LIBERO-Plus & 4 & \code{libero\_spatial}, \code{libero\_object}, \code{libero\_goal}, \code{libero\_10} \\
\rev{LIBERO-Pro} & \rev{16} & \rev{\code{libero\_\{goal,spatial,10,object\}\_\{object,swap,lan,task\}}} \\
RoboTwin 2.0 & 2 & \code{demo\_clean} (Easy), \code{demo\_randomized} (Hard) \\
\rev{RoboDojo} & \rev{2} & \rev{\code{arx\_x5}, \code{arx\_x5\_randomized}} \\
CALVIN & 3 & \code{task\_D\_D}, \code{task\_ABC\_D}, \code{task\_ABCD\_D} \\
RoboCasa GR-1 Tabletop & 1 & \code{tabletop} \\
MetaWorld & 1 & \code{mt50} \\
\rev{RLBench} & \rev{1} & \rev{\code{peract\_18}} \\
\rev{robomimic} & \rev{1} & \rev{\code{ph\_low\_dim}} \\
\bottomrule
\end{tabular}
\end{table}

\paragraph{Benchmark adapters.}
An adapter implements what differs between benchmarks: building work items from the suite's tasks
and episodes, starting a native rollout process for an item, translating the benchmark's observations
into named features and the Policy's actions into the benchmark's convention, and parsing native
results into episode records. Task initialization, reset, action application, termination, and scoring run in the
benchmark's native evaluator, which the adapter instruments to route inference to a Policy replica
and to emit episode records. \rev{The catalog covers ten benchmarks, whose suites are listed in Table~\ref{tab:v2-compatibility}: LIBERO,
LIBERO-Plus, LIBERO-Pro \citep{liberopro2025}, RoboTwin 2.0, RoboDojo \citep{robodojo2026}, CALVIN, RoboCasa GR-1
Tabletop, MetaWorld MT50 \citep{metaworld2019}, RLBench with the 18 PerAct tasks \citep{rlbench2020,peract2022},
and robomimic \citep{robomimic2021}. The catalog is extensible: a further benchmark is supported by adding its
adapter, a manifest of its suites, and a script that builds its environment, and every Policy whose required
features the benchmark provides then runs on it unchanged.} The measurements in this report use LIBERO-Long and
RoboTwin Easy, and the cross-instance run uses MetaWorld MT50.

\paragraph{Adding a Policy.}
A new Policy is one class. \rev{The class subclasses \code{Policy} and implements \code{load} and \code{act}, while \code{reset}, \code{end\_episode}, and \code{unload} default to no-ops.}
\code{load} returns the Policy's capabilities, including required features, chunk lengths, action
representation, and observation strides. The class's script ends with \code{run(policy)}, which serves it on the socket the Control Runtime provides. The shared Control Runtime and every
benchmark adapter are reused unchanged, and the same class serves every benchmark whose features it
declares.

\rev{Appendix~\ref{app:v2-policy-examples} gives two condensed Policies: a stateless StarVLA Policy that
declares an observation stride equal to its chunk length, and a stateful X-VLA Policy that keys its
per-episode state by episode identifier.}

\section{Evaluation}
\label{sec:v2-evaluation}
Our evaluation answers the following questions:
\begin{itemize}
  \item Does \system{} preserve benchmark outcomes (\S\ref{sec:v2-reference})?
  \item How does its evaluation time compare with existing systems (\S\ref{sec:v2-throughput})?
  \item How much does each design component contribute (\S\ref{sec:v2-cost})?
  \item How does it scale with GPUs, workers, and Policy replicas (\S\ref{sec:v2-scalability})?
\end{itemize}

\subsection{Experimental Setup}
\label{sec:v2-methodology}
\paragraph{Workloads and checkpoints.}
The cross-system comparison uses LIBERO-Long (suite \code{libero\_10}), 10 tasks with
50 initial states each and seed 7, evaluated with the public StarVLA checkpoint
\nolinkurl{StarVLA/Qwen2.5-VL-OFT-LIBERO-4in1} \citep{starvla2026,qwen25vl2025}. The ablation study and a
second cross-system comparison on one RTX 4090 use RoboTwin Easy (suite \code{demo\_clean}), 50 tasks with 10 episodes each, evaluated with
\nolinkurl{StarVLA/Qwen3-VL-OFT-Robotwin2} \citep{qwen3vl2025}, whose chunk length is 16. Each workload
has a frozen ordered manifest of 500 episode keys, and every run of that workload receives
the same keys (Appendix~\ref{app:v2-setup}). \rev{The comparison with the official
evaluation on four RTX PRO 5000 GPUs also runs the standard RoboTwin evaluation protocol on RoboTwin Easy, 100 episodes per task and 5,000
in all.}

\paragraph{Baselines.}
On LIBERO-Long, we compare \system{} with three evaluation systems that run the same checkpoint and manifest: the official
StarVLA evaluation (its policy server and LIBERO evaluator), vla-eval \citep{vlaeval2026}, and RLinf
\citep{rlinf2026}. RLinf was tuned once offline for the host, which took 1,704\,s. vla-eval runs
the fastest configuration found by 9,780\,s of offline tuning over its shard count, model-server replicas, maximum
batch size, and batch wait (Table~\ref{tab:v2-baseline-tuning}): 24 shards share one server that
batches up to 24 observations with a 5\,ms wait. \rev{\textbf{RoboScope Kit (Search)} runs \system{} with autotuning (Section~\ref{sec:v2-selection}) and a StarVLA Policy that uses the same camera-image preprocessing and gripper processing as the StarVLA client. The search selects $K12/g3$ in all three of its launches, and the cross-system comparison evaluates this configuration with CUDA MPS off, the default.} \rev{\textbf{RoboScope Kit (Search+MPS)} runs the same search and evaluation with NVIDIA CUDA MPS on (\code{--cuda-mps}), which lets the replicas run their kernels concurrently on the GPU instead of in time slices. In this setting, the search selects $K16/g4$ in all three launches (Section~\ref{sec:v2-cost}).} The ablation study compares the configurations chosen by the search with fixed configurations, in which $K$ and $g$ are set by hand and no search runs.
On RoboTwin Easy, we compare \system{} with the same three systems on the same checkpoint and 500 episodes, every
system taking the episode instruction from the frozen manifest. The official StarVLA evaluation runs its
RoboTwin evaluator with one environment and one model. RLinf runs 8 environments in 8 processes and vla-eval 9
shards on one server at batch size~1, the configurations their tuning selected. RoboScope Kit (Search) selects
$K8/g1$, with CUDA MPS off (Appendix~\ref{app:v2-robotwin-baselines}).

\paragraph{Hardware.}
\rev{The GPUs are 48\,GiB RTX 4090 GPUs and, for the comparison with the official evaluation on four GPUs, 72\,GiB RTX
PRO 5000 GPUs. Appendix~\ref{app:v2-hosts} gives the host, CPU quota, and number of launches of every experiment.}

\paragraph{Metrics.}
Success rate is the fraction of successful episodes. \rev{Outcome agreement between two runs is the fraction of episode keys with the same success outcome, and terminal-step agreement also requires the same terminal step.} \rev{Evaluation time covers the formal evaluation and excludes process initialization, autotuning or offline tuning, and cleanup. End-to-end time is the wall time of the launch without the search or tuning.} \rev{Throughput is policy-controlled environment steps per second of evaluation time, and speedup is the ratio of reference to candidate evaluation time. Where both systems ran three launches, the speedup carries a 95\% $t$ interval (Appendix~\ref{app:v2-performance}), and single launches carry none.} On LIBERO-Long, $\Delta$ success carries a 95\% paired
task-cluster bootstrap interval (10 tasks, 20,000 draws). Outcomes are equivalent when the $\Delta$ interval lies within $\pm2$
percentage points.

\subsection{Outcome Agreement}
\label{sec:v2-reference}
\begin{figure}[tbp]
\centering
\includegraphics[width=\linewidth]{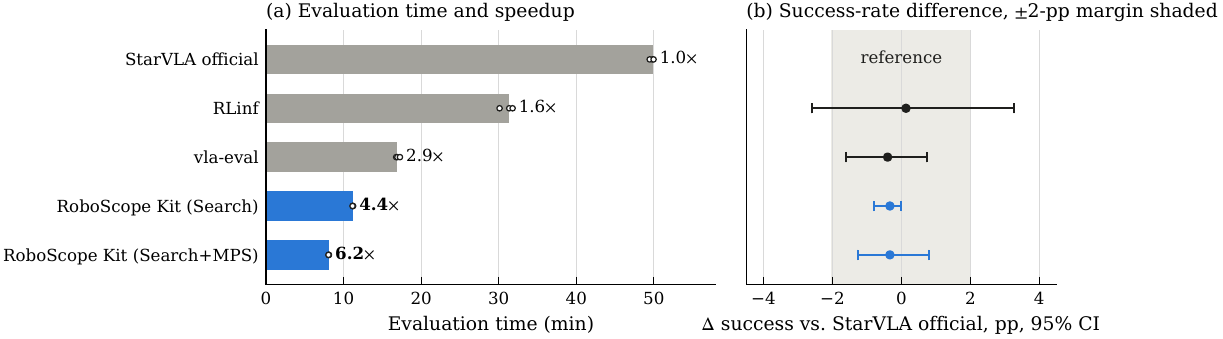}
\caption{LIBERO-Long on one RTX 4090 with RoboScope Kit (Search), which selects $K12/g3$, and RoboScope Kit
(Search+MPS), which selects $K16/g4$ under CUDA MPS. \rev{(a) Bars show each system's median evaluation time and circles show its launches, and labels give the speedup over the official StarVLA evaluation.} \rev{(b) Success rate minus that of the official StarVLA evaluation, with a 95\% paired task-cluster bootstrap interval. The shaded band marks the $\pm2$-point equivalence margin.}
Table~\ref{tab:v2-c17-comparison} lists the values.}
\label{fig:v2-w1-results}
\end{figure}

\paragraph{Agreement with the official evaluation.}
A fixed $K1/g1$ \system{} run from the worker and replica sweep matches each of the three official
launches on all 500 LIBERO-Long success outcomes (Appendix~\ref{app:v2-outcome}). \rev{In the cross-system comparison, the official evaluation scores 90.2\% and RoboScope Kit (Search) 89.9\%, $-0.3$ percentage points (95\% interval $-0.8$ to $0.0$), with 97.9\% of episodes having the official outcome. RoboScope Kit (Search+MPS) also scores 89.9\%, $-0.3$ points ($-1.3$ to $+0.8$), with 96.5\% agreement (Figure~\ref{fig:v2-w1-results}).} Under the $\pm2$-point
margin, both are equivalent to the official evaluation. On RoboTwin Easy, the stride-16 run on four RTX 4090 GPUs matches each of the three
$K1/g1$ runs on one RTX 4090 on every success outcome and terminal step, and the stride-1 runs on four RTX 4090 GPUs
at $K4/g4$, $K16/g4$, and $K32/g4$ agree with one another. These stride-1 runs score 235/500 against 231/500 for the
$K1/g1$ runs, a difference of $+0.8$ percentage points (95\% interval $-2.0$ to $+3.8$), with
89.2\% of episodes having the same outcome.

\paragraph{Agreement with other evaluation systems.}
RLinf scores 90.3\%, $+0.1$ percentage points relative to the official evaluation [95\% interval $-2.6$ to
$+3.3$], and shares the official outcome on 88.1\% of episodes. vla-eval scores 89.8\%, 0.4 points below
the official evaluation [95\% interval $-1.6$ to $+0.7$], with 91.7\% agreement. For comparison, repeated
launches of the same system agree on 86.2--88.4\% of episodes for RLinf, 93.6--95.0\% for vla-eval,
97.8--99.0\% for RoboScope Kit (Search), 97.4--98.4\% for RoboScope Kit (Search+MPS), and 100\% for the
official evaluation, so RLinf differs from the official evaluation about as much as from its own
repeated launches. RLinf samples its actions during evaluation (Appendix~\ref{app:v2-rlinf-sampling}).

\paragraph{Agreement on RoboTwin Easy.}
On the 500 RoboTwin Easy episodes on one RTX 4090, the official StarVLA evaluation and vla-eval reproduce
RoboScope Kit (Search) episode for episode: all three score 232/500, with the same success outcome and the same
terminal step in every episode (Table~\ref{tab:v2-robotwin-comparison}). \rev{RLinf, which evaluates the mean action here, scores 230/500 and shares the outcome on 434 episodes (86.8\%) and the terminal step on 287 (57.4\%). In 32 episodes only RLinf succeeds, and in 34 only \system{} succeeds.} RLinf batches inference over its environments
and executes actions with the official control loop, \rev{in which \system{} hoists the duplicate passive force}
(Appendix~\ref{app:v2-robotwin-baselines}).

\paragraph{Sources of outcome differences.}
Within \system{}, LIBERO-Long runs with 2 to 20 simulator workers differ from a one-worker run on 1 to
28 of 500 success outcomes, while the success count moves by at most ten (Table~\ref{tab:v2-h1-scaling}).
The differences follow task assignment: they fall in tasks split across workers.
Appendix~\ref{app:v2-outcome} breaks them down by episode and condition.

\subsection{Performance}
\label{sec:v2-throughput}
\paragraph{Evaluation-time comparison.}
Appendix~\ref{app:v2-performance} lists every launch with its initialization, search, and tuning
time.

\paragraph{Results.}
RoboScope Kit (Search) completes LIBERO-Long in a median 671\,s of evaluation time, against 2,990\,s for the
official StarVLA evaluation, 1,882\,s for RLinf, and 1,016\,s for vla-eval: 4.4$\times$ (95\% CI
[4.4, 4.5]) faster than StarVLA, 2.8$\times$ ([2.6, 3.0]) faster than RLinf, and 1.5$\times$
([1.5, 1.6]) faster than vla-eval at its tuned configuration. Its throughput, 204.6 steps/s, exceeds
vla-eval's 133.5, RLinf's 74.2, and StarVLA's 45.8, with outcomes equivalent to the official evaluation.
RoboScope Kit (Search+MPS) completes LIBERO-Long in a median 485\,s at
282.5 steps/s: 6.2$\times$ ([6.1, 6.2]) faster than StarVLA, \rev{3.8$\times$} ([3.6, 4.1]) faster
than RLinf, 2.1$\times$ ([2.0, 2.2]) faster than vla-eval, and 1.4$\times$ ([1.37, 1.40]) faster
than RoboScope Kit (Search) on the same host and cores.

\paragraph{RoboTwin Easy.}
On one RTX 4090, RoboScope Kit (Search) completes the 500 RoboTwin Easy episodes in 1,916\,s of evaluation time, against
25,238\,s for the official StarVLA evaluation, 14,072\,s for vla-eval, and 4,237\,s for RLinf: 13.2$\times$,
7.3$\times$, and 2.2$\times$ faster (Table~\ref{tab:v2-robotwin-comparison}). Its throughput, 110.2 steps/s,
exceeds RLinf's 49.9, vla-eval's 15.0, and the official evaluation's 8.4.
GPU utilization does not order the systems:
vla-eval keeps the GPU 85.1\% busy but is the second slowest, because each of its nine shards renders every step
in its own engine and renderer, while \system{} renders once per 16-step chunk in a shared one at 63.5\%.
The official evaluation runs one episode at a time and also \rev{runs the expert check} for every seed, which takes
3,177\,s of its evaluation time.

\begin{table}[tbp]
\centering
\caption{Speed and outcome agreement in the RoboTwin Easy cross-system comparison.}

\label{tab:v2-robotwin-comparison}
\footnotesize
\setlength{\tabcolsep}{4pt}
\begin{tabular}{@{}l l r r r r r r r r@{}}
\toprule
System & Configuration & Successes & \shortstack{Same\\outcome} & \shortstack{Same\\step} & \shortstack{Eval.\\time (s)} & \shortstack{End-to-\\end (s)} & Steps/s & Slowdown & \shortstack{GPU\\util.} \\
\midrule
StarVLA official & 1 environment & 232 & 500 & 500 & 25,238 & 27,165 & 8.4 & 13.2$\times$ & 38.1\% \\
vla-eval & 9 shards, batch 1 & 232 & 500 & 500 & 14,072 & 14,085 & 15.0 & 7.3$\times$ & 85.1\% \\
RLinf & 8 environments & 230 & 434 & 287 & 4,237 & 4,286 & 49.9 & 2.2$\times$ & 23.8\% \\
RoboScope Kit (Search) & $K8/g1$ & 232 & -- & -- & 1,916 & 1,931 & 110.2 & 1.0$\times$ & 63.5\% \\
\bottomrule
\end{tabular}
\par\smallskip
{\raggedright All four systems evaluate the same 500 episodes with the same checkpoint on one RTX 4090, one launch
each. \rev{RoboScope Kit reuses the Engine and Renderer and hoists the duplicate passive force (Section~\ref{sec:v2-runtime}).} Same outcome and same step count the episodes whose success outcome or terminal step, respectively, matches
RoboScope Kit (Search). \rev{Evaluation time covers the formal evaluation (Section~\ref{sec:v2-methodology}), including the official evaluation's expert check. End-to-end time is the wall time of the launch apart from search or tuning.} Steps/s is policy steps per second of evaluation time, and slowdown is the row's evaluation time divided
by that of RoboScope Kit (Search). GPU utilization is the mean of each system's own sampler, or of a 1\,Hz host
trace for the official evaluation. Appendix~\ref{app:v2-robotwin-baselines} gives the details.\par}

\end{table}

\paragraph{\rev{RoboTwin Easy on four RTX PRO 5000 GPUs.}}
\rev{On one host with four RTX PRO 5000 GPUs, we compare \system{} with the official StarVLA evaluation, which
runs one evaluator per GPU, by evaluation time as defined in Table~\ref{tab:v2-4gpu-headline}. The standard
RoboTwin Easy protocol is 100 episodes for each of the 50 tasks, and the official evaluation completes these 5,000 episodes
in 63,036\,s, 17.5 hours. \system{} runs $K60/g8$, the configuration its autotuner selects on the 500 episodes below, and takes 3,985\,s: 15.8$\times$ faster. It matches the official evaluation on all 5,000 success outcomes and terminal steps,
with 2,313 successes (46.3\%). \system{} reads its seeds from a cache built beforehand by the official expert
check, whereas the StarVLA evaluator repeats that check before every episode, which takes 9.1\% of its episode time.}

\rev{On the 500 RoboTwin Easy episodes, the official evaluation takes a mean 7,076\,s over three launches. The
autotuner selects $K60/g8$, and in one launch \system{} evaluates them in 706\,s, 10.0$\times$ faster,
matching the official evaluation on every success outcome and terminal step.}

{\revcolor
\begin{table}[tbp]
\revcolor
\centering
\small
\caption{RoboTwin Easy on four RTX PRO 5000 GPUs against the official StarVLA evaluation on the same GPUs. Evaluation
time, the mean of the runs, runs for the official evaluation from the start of the first episode to the end of the last, and
for \system{} from the end of replica loading to the end of the last episode. Every \system{}
launch matches the official evaluation on every success outcome and terminal step.}
\label{tab:v2-4gpu-headline}
\begin{tabular}{@{}l r r r r@{}}
\toprule
System & Runs & Successes & Eval.\ time (s) & vs.\ official \\
\midrule
\multicolumn{5}{@{}l}{\emph{Standard protocol, 5,000 episodes (50 tasks $\times$ 100)}} \\
StarVLA official & 1 & 2,313 & 63,036 & 1.0$\times$ \\
\system{}, $K60/g8$$^{*\dagger}$ & 1 & 2,313 & 3,985 & 15.8$\times$ \\
\addlinespace
\multicolumn{5}{@{}l}{\emph{500 episodes (50 tasks $\times$ 10)}} \\
StarVLA official & 3 & 230 & 7,076 & 1.0$\times$ \\
\system{}, $K60/g8$$^{\dagger}$ & 1 & 230 & 706 & 10.0$\times$ \\
\bottomrule
\end{tabular}
\par\smallskip
\raggedright\footnotesize \rev{$^{*}$Engine and Renderer reuse, the hoisted passive force, one-episode work items, and parallel replica start across GPUs. The 5,000-episode launch also frees each finished task's scene and renders the three cameras the Policy reads (Appendix~\ref{app:v2-ablation}).}
$^{\dagger}$Selected by the autotuner on the 500 episodes (Appendix~\ref{app:v2-ablation}).\par
\end{table}
}

\subsection{Ablation Study}
\label{sec:v2-cost}
\paragraph{Separating simulation and inference.}
Separation lets the Control Runtime add a Policy replica without adding simulators. On one RTX 4090, going from
$K12/g1$ to $K12/g2$ shortens LIBERO-Long evaluation from 1,216\,s to 748\,s (1.6$\times$) and
raises mean GPU utilization from 46.3\% to 87.7\% (Table~\ref{tab:v2-h1-scaling}). At $K12/g2$, peak GPU
memory is 25.0\,GiB.

\paragraph{Autotuning.}
On one RTX 4090 with a 20-core quota, the diagonal search
selects $K12/g3$ in all three launches
(Tables~\ref{tab:v2-h1-scaling} and~\ref{tab:v2-search-launches}). The worker phase reaches the knee at $K4/g1$:
doubling to $K8/g1$ gains less than 25\%. The replica phase then accepts $K8/g2$ and $K12/g3$ and rejects
$K16/g4$, which gains only 2.1--2.4\%, under the 3\% threshold. The selected configuration evaluates LIBERO-Long
1.1$\times$ faster than $K12/g2$, the fastest fixed configuration with at most two replicas, and
5.3$\times$ faster than fixed $K1/g1$. Its steady-state probes
of $K12/g3$ measure within 1.2\% of the formal throughput of the same launch.
Figure~\ref{fig:v2-scaling}a compares the diagonal search with a \emph{grid search}, which runs every $K/g$
configuration as a full evaluation and keeps the fastest: the grid search takes eighteen full evaluations, the
diagonal search seven short probes. The diagonal search's choice evaluates within 1.2\% of the fastest grid
configuration, $K16/g4$ (670.9\,s), inside the 3\% gain below which it stops adding replicas.

\paragraph{Autotuning under CUDA MPS.}
With CUDA MPS on and 12 pinned cores, a further replica keeps paying off
(Figure~\ref{fig:v2-scaling}b): the diagonal search accepts $K16/g4$, which adds 8--9\% over $K12/g3$, and
stops because $K20/g5$ is predicted to exceed GPU memory. \rev{It selects $K16/g4$ in all three launches. These launches form the RoboScope Kit (Search+MPS) row of Section~\ref{sec:v2-throughput} and evaluate 1.4$\times$ faster than $K12/g3$ with MPS off on the same host and cores.} A grid of eighteen
fixed configurations under MPS, $g$ from 1 to 4
(Table~\ref{tab:v2-mps-grid}), confirms the choice: $K16/g4$ is its fastest configuration at 482.5\,s,
ahead of $K20/g4$ (483.2\,s) and $K12/g4$ (489.2\,s), and every configuration with fewer replicas
is at least 5.9\% slower.

\begin{figure}[tbp]
\centering
\includegraphics[width=\linewidth]{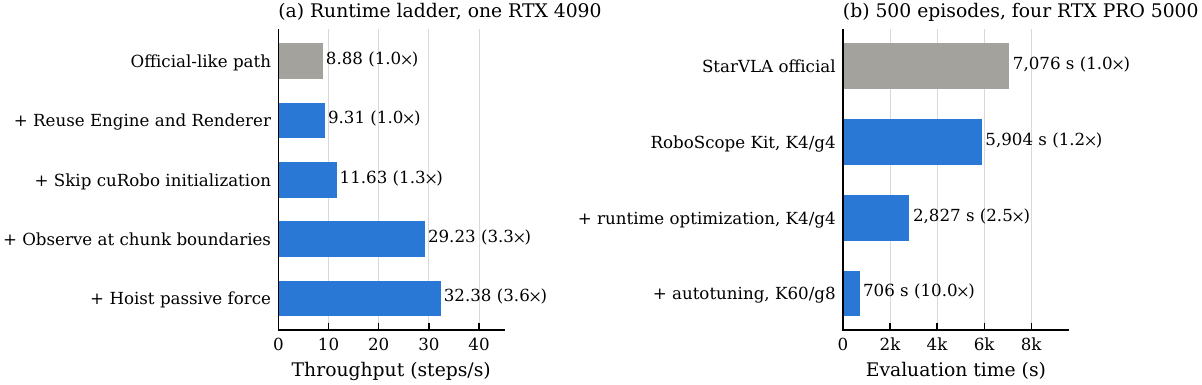}
\caption{RoboTwin Easy ablations. \rev{(a) Throughput of the Simulation Runtime ladder on one RTX 4090, with a Policy that holds the current joint state in place of a model. Each row adds one change to the row above \rev{and runs once, on one task with three seeds}.} \rev{(b) Evaluation time of the 500 episodes with the real StarVLA Policy on four RTX
PRO 5000 GPUs, from the official evaluation along the main measured \system{} configurations: $K4/g4$
without runtime optimization, then runtime optimization, and $K60/g8$ selected by the autotuner. The first row is the mean of
three launches, and the others ran once.} Labels give each value
and, in parentheses, the cumulative speedup over the first row.
Tables~\ref{tab:v2-e18-ladder}, \rev{\ref{tab:v2-speedup-chain}, and~\ref{tab:v2-4gpu-headline}} list the values.}
\label{fig:v2-ablation}
\end{figure}

\paragraph{Runtime optimization.}
A controlled ladder on one RTX 4090 adds the RoboTwin Simulation Runtime changes one at a time, with a Policy
that holds the current state (Figure~\ref{fig:v2-ablation}a). Engine and Renderer reuse raises
throughput from 8.88 to 9.31 steps/s. For the joint-position Policy, skipping cuRobo initialization
reduced total setup time for three episodes from 45.2\,s to 1.4\,s and raised throughput from 9.31
to 11.63 steps/s. Observing at 16-step chunk boundaries
reduces rendered observations from 2,100 to 132 and raises throughput 2.5$\times$, to 29.23 steps/s,
and hoisting the duplicate passive force reaches 32.38 steps/s, 3.6$\times$ the official-like path.
With the real StarVLA Policy at $K16/g4$ on four RTX 4090 GPUs, stride 16 with its run setup is 4.6$\times$ faster than
stride 1 (Tables~\ref{tab:v2-4090-chain} and~\ref{tab:v2-stride16}). \rev{On four RTX PRO 5000 GPUs at
$K4/g4$, with the real StarVLA Policy and 100 episodes, the first two of each of the 50 tasks, Engine and Renderer reuse, the
\rev{skipping cuRobo initialization, hoisting the duplicate passive force,} and chunk-boundary observation together cut evaluation time from
1,349\,s to 552\,s, 2.4$\times$, and every episode keeps the official outcome and terminal step
(Table~\ref{tab:v2-pro5000-ladder}).} \rev{\rev{RoboTwin renders four cameras for every observation, and the Policy reads three of them. \system{} renders those three, with images bit-identical to the official ones (Appendix~\ref{app:v2-ablation}).}}

\paragraph{\rev{From the official evaluation to $K60/g8$ on four RTX PRO 5000 GPUs.}}
\rev{Figure~\ref{fig:v2-ablation}b steps through the main measured configurations
on the 500 RoboTwin Easy episodes on four RTX PRO 5000 GPUs, by evaluation time, from the official evaluation (7,076\,s) to
$K60/g8$, the configuration the autotuner selects (706\,s, Table~\ref{tab:v2-4gpu-headline}).
\system{} at $K4/g4$ with runtime optimization off is 1.2$\times$ faster than the
official evaluation, and runtime optimization adds 2.1$\times$. The autotuner then adds workers and replicas,
with one-episode work items and parallel replica start, for another 4.0$\times$ and 10.0$\times$ in total.
The search maximizes steady-state throughput, which pays off most on long runs such as the 5,000 episodes.
Table~\ref{tab:v2-speedup-chain} lists the intermediate configurations.}

\subsection{Scalability}
\label{sec:v2-scalability}
\begin{figure}[tbp]
\centering
\includegraphics[width=\linewidth]{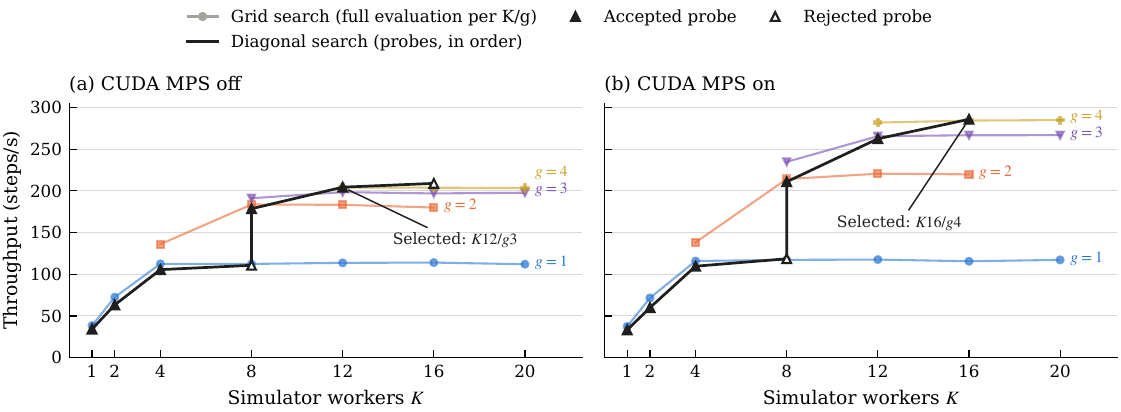}
\caption{Grid search and diagonal search on LIBERO-Long, one RTX 4090. \rev{Grid-search lines are full evaluations at fixed $K/g$, one launch per configuration, one line per replica count $g$, labelled at its end. The diagonal-search path joins the steady-state probes in the order the search ran them, mean of three launches, with $g$ marked where no grid line has that replica count. Filled triangles are probes the search accepted and hollow triangles probes it rejected, and a leader line points to the configuration the search selects.} (a) CUDA MPS off, with a 20-core quota:
the grid search takes eighteen launches, the diagonal search seven probes and selects $K12/g3$, within 1.2\% of the fastest grid configuration, $K16/g4$. \rev{(b) CUDA MPS on, with 12 pinned cores: the grid search takes eighteen launches, the diagonal search seven probes and selects $K16/g4$, the fastest grid configuration. The diagonal search skips $K20/g5$ because its predicted memory exceeds the GPU.} GPU utilization of the grid configurations is in
Tables~\ref{tab:v2-h1-scaling} and~\ref{tab:v2-mps-grid}.}
\label{fig:v2-scaling}
\end{figure}

\paragraph{Worker and Policy-replica scaling.}
Figure~\ref{fig:v2-scaling}a and Table~\ref{tab:v2-h1-scaling} sweep $K$ and $g$ on one RTX 4090 with a
20-core quota. With one
replica, throughput rises from 38.4 steps/s at $K1$ to 112.5 at $K4$ and stays at 112--114 up to
$K20$, while GPU utilization stays between 45.6\% and 46.8\%: the replica is the bottleneck. \rev{A second replica moves saturation to $K8$, raises throughput to 183.6 steps/s and utilization to 86.7\%, and reaches 4.8$\times$ over $K1/g1$ at $K12/g2$. A third and a fourth replica add 8\% and a further 3\%, to 5.3$\times$ at $K16/g4$.} Per-worker efficiency falls from 0.94 at $K2$ to 0.15 at $K20$
with one replica. Success-outcome agreement with $K1/g1$ is 99.8\% at $K2$ and 94.4--98.8\% at larger
$K$, and terminal-step agreement falls from 98.4\% to 41.8\% as more workers share each task's episodes.

\begin{figure}[tbp]
\centering
\includegraphics[width=\linewidth]{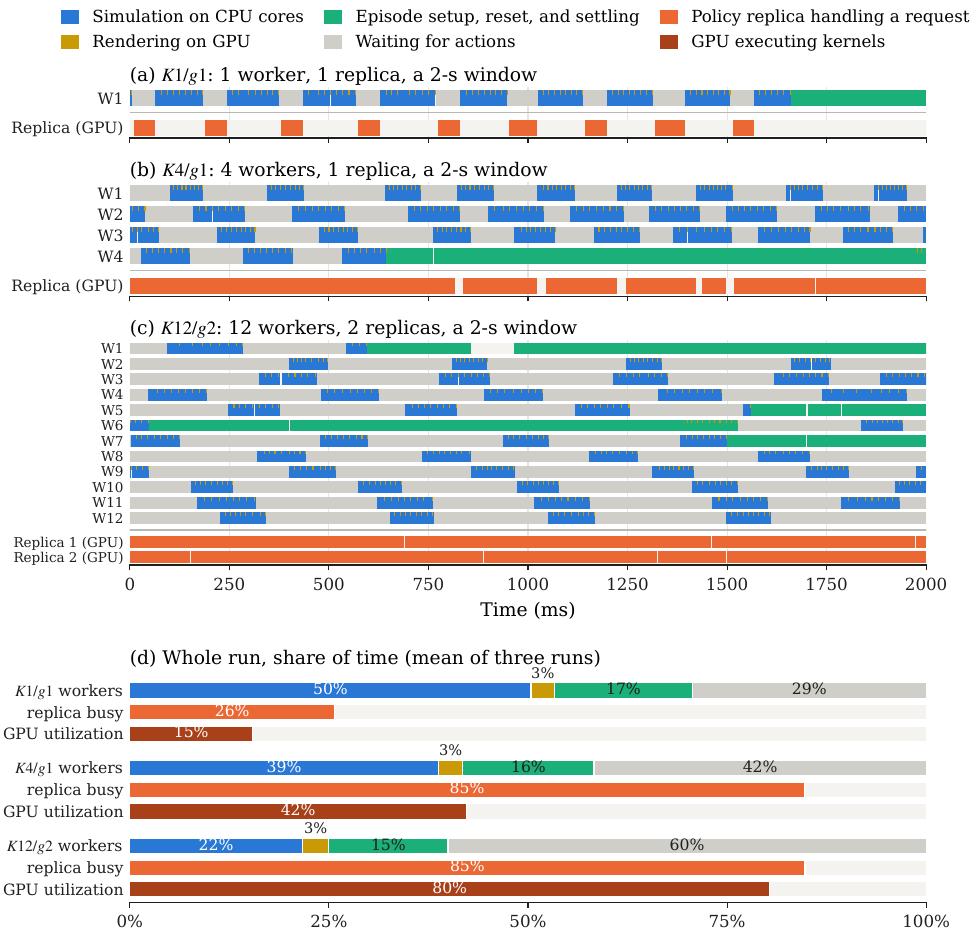}
\caption{Execution timeline of \system{} simulator workers (W) and Policy replicas on the 50 LIBERO-Long
episodes of Figure~\ref{fig:v2-native-timeline}, one RTX 4090, with the categories of that figure.
(a)--(c) A 2-s window of one run per configuration. \rev{Rendering runs on the GPU at the end of each simulation step and is drawn there as a tick. Its time is counted in the blue bars, and (d) shows it as its own share.} (d) Worker time, replica busy time, and GPU utilization over the whole run, mean of three runs.}
\label{fig:v2-kit-timeline}
\end{figure}

\paragraph{Where the bottleneck moves.}
A stage-timed profile of 50 LIBERO-Long episodes shows why (Figure~\ref{fig:v2-kit-timeline},
Table~\ref{tab:v2-time-profile}). At $K4/g1$, the
workers saturate the replica: it is busy 85\% of the time, requests wait 48\,ms for it, and workers
spend 42\% of their time waiting for actions. $K12/g2$ overlaps these
launch-bound inferences and raises GPU utilization from 42\% to 80\%. At $K12/g2$,
inference again sets the pace: requests wait 221\,ms, and each inference stretches to
84\,ms under contention. Adding simulators moves the bottleneck to inference, and adding replicas
moves it to the GPU, so the right numbers of workers and replicas depend on each other and on the
workload and host.

\paragraph{GPU scaling.}
\rev{At $K4/g4$, four RTX 4090 GPUs evaluate the 500 RoboTwin Easy episodes in 5,934\,s against 21,582\,s on one (three runs each, Table~\ref{tab:v2-4090-chain}). The speedup is 3.6$\times$, and the parallel efficiency is 0.91.} \rev{On four RTX PRO 5000 GPUs, throughput at $K32/g4$
is 4.0$\times$ that of one GPU at $K8/g1$
(Table~\ref{tab:v2-pro5000-scaling}).}

\paragraph{Cross-instance execution.}
On our evaluation platform, four concurrent instances complete the full MetaWorld MT50 protocol,
2,500 episodes, and the platform merges their shard reports into one result.

\section{Related Work}
\label{sec:v2-related}
\paragraph{Benchmarks and evaluation scope.}
Manipulation benchmarks differ in what a successful episode establishes. LIBERO organizes tasks to
study knowledge transfer \citep{libero2023}, and LIBERO-Plus tests robustness under systematic
perturbations \citep{liberoplus2025}. \rev{CALVIN evaluates sequences of language-conditioned tasks \citep{calvin2022}, and RoboTwin emphasizes bimanual manipulation and scalable task generation \citep{robotwin2025,robotwin22025}. RoboCasa provides diverse household scenes and tasks \citep{robocasa2024}, including the GR-1 Tabletop collection released with GR00T N1 \citep{gr00t2025}.}

\paragraph{Evaluation pipelines and reproducibility.}
Native evaluators couple task selection and reset with model inference, rollout termination, and
score aggregation. OpenVLA supplies benchmark evaluation scripts, and StarVLA unifies evaluation
interfaces within its model-development codebase \citep{openvla2024,starvla2026}. vla-eval evaluates vision-language-action models across benchmarks with batch-parallel
evaluation \citep{vlaeval2026}, and RLinf, an infrastructure for reinforcement learning of embodied
models, runs parallel environment evaluation \citep{rlinf2026}. \rev{XPolicyLab specifies common observation, action, and trajectory schemas with a minimal policy adapter and separates policy inference from environment execution in a client/server architecture, so that $N$ policies and $M$ environments need $O(N+M)$ integrations \citep{xpolicylab2026}. RoboDojo evaluates such policies with heterogeneous parallel simulation in Isaac Sim \citep{robodojo2026}.} \system{} shares the integrate-once goal and
adds what evaluation speed needs: separately sized pools of simulator workers and Policy replicas,
measurement-based autotuning, and runtime optimization declared through the Policy's observation
stride. SimplerEnv relates simulated to
real-robot performance \citep{simplerenv2025}. \rev{Gymnasium standardizes the environment boundary \citep{gymnasium2025}, while \system{} standardizes the complementary model boundary and keeps benchmark-specific rollout logic in adapters.} Findings on the sensitivity of robot imitation learning
\citep{robomimic2021} and the uncertainty of few-run reinforcement-learning evaluation
\citep{henderson2018,rliable2021} motivate \system{}'s frozen manifests, archived configurations, paired
episode records, and separate reporting of evaluation and search costs.

\paragraph{Simulation and execution efficiency.}
MuJoCo, robosuite, and SAPIEN supply physics and task infrastructure for manipulation
\citep{mujoco2012,robosuite2020,sapien2020}. \rev{Isaac Gym parallelizes simulation on the GPU, and ManiSkill3 parallelizes both simulation and rendering on the GPU \citep{isaacgym2021,maniskill32024}. EnvPool accelerates supported environments with a parallel execution engine \citep{envpool2022}.} \system{} coordinates separate
simulator workers and Policy replicas around existing benchmark runtimes, which complements these
engines. Distributed reinforcement-learning systems separate environment stepping from learning
\citep{impala2018}, and SEED RL and Sample Factory move inference into batched accelerator workers
\citep{seedrl2020,samplefactory2020}. \system{} applies this separation to evaluation, where the
checkpoint is fixed and the objective is complete, reproducible episode records rather than sample
throughput for learning. Its chunk-boundary observation removes rendering that an action-chunk Policy
never reads.

\paragraph{Policies and the model--benchmark contract.}
Generalist and vision-language-action policies vary in sensory inputs, action representations, and
inference procedures \citep{openvla2024,octo2024,pi02025}. Diffusion Policy and ACT predict action
sequences executed with receding-horizon or chunked control \citep{diffusionpolicy2025,act2023}, and
OpenVLA-OFT studies speed and success under changes to action decoding and adaptation
\citep{oft2025}. These choices determine when a policy needs observations, what state it keeps across
steps, and how its outputs must be interpreted. \system{}'s Policy boundary therefore keeps
checkpoint-specific processing with the model, declares chunk lengths and observation strides, and
keys state by episode.

\section{Discussion}
\label{sec:v2-discussion}
\rev{The autotuner runs before each evaluation, and its time is reported in Appendix~\ref{app:v2-search-time}.} \rev{Reusing the selected configuration for the
same host, Policy, and benchmark removes the search from repeated evaluations: with a run cache, a second launch reads the
stored configuration and runs no probes (Appendix~\ref{app:v2-ablation}).}

\section{Conclusion}
Profiling the StarVLA evaluators shows that simulation occupies CPU cores while one loaded model
leaves the GPU mostly idle, and that an action-chunk Policy reads only one of every 16 RoboTwin
renders. \rev{\system{} integrates each model once as a Policy and each benchmark once as an adapter around its native evaluator, and at $K1/g1$ it matches all 500 official LIBERO-Long outcomes.}
Scheduling separates simulators from Policy replicas and sizes both by measurement: on one RTX 4090, $K12/g2$
reaches 4.8$\times$ the throughput of $K1/g1$,
and with $K12/g3$, selected by the autotuner, LIBERO-Long
evaluation runs 4.4$\times$ faster than the official evaluation with equivalent outcomes, or
6.2$\times$ with CUDA MPS, where the search selects $K16/g4$. \rev{On four RTX PRO 5000 GPUs,
runtime optimization, which renders only the observations a chunked Policy reads, and $K60/g8$, the configuration
the autotuner selects, evaluate the 5,000 episodes of the standard RoboTwin Easy protocol 15.8$\times$ faster than the
official evaluation on the same GPUs, matching it on every success outcome and terminal step.} Deterministic sharding extends an evaluation across
GPU instances.

\bibliographystyle{plainnat-shortauthors}
\bibliography{references,references-extra}

@article{vlaeval2026,
  title   = {vla-eval: A Unified Evaluation Harness for Vision-Language-Action Models},
  author  = {Choi, Suhwan and Lee, Yunsung and Park, Yubeen and Kim, Chris Dongjoo and Krishna, Ranjay and Fox, Dieter and Yu, Youngjae},
  journal = {arXiv preprint arXiv:2603.13966},
  year    = {2026},
  url     = {https://github.com/allenai/vla-evaluation-harness}
}

@inproceedings{rlinf2026,
  title     = {{RLinf}: Flexible and Efficient {Large-Scale} Reinforcement Learning via {Macro-to-Micro} Flow Transformation},
  author    = {Yu, Chao and Wang, Yuanqing and Guo, Zhen and Lin, Hao and Xu, Si and Zang, Hongzhi and Zhang, Quanlu and Wu, Yongji and Zhu, Chunyang and Hu, Junhao and Huang, Zixiao and Wei, Mingjie and Xie, Yuqing and Yang, Ke and Dai, Bo and Xu, Zhexuan and Du, Jiakun and Wang, Xiangyuan and Fu, Xu and Shi, Letong and Liu, Zhihao and Chen, Kang and Liu, Weilin and Liu, Gang and Li, Boxun and Yang, Jianlei and Yang, Zhi and Dai, Guohao and Wang, Yu},
  booktitle = {20th USENIX Symposium on Operating Systems Design and Implementation (OSDI 26)},
  pages     = {829--846},
  publisher = {USENIX Association},
  address   = {Seattle, WA},
  month     = jul,
  year      = {2026},
  url       = {https://www.usenix.org/conference/osdi26/presentation/yu-chao}
}

@inproceedings{xvla2026,
  title     = {{X-VLA}: Soft-Prompted Transformer as Scalable Cross-Embodiment Vision-Language-Action Model},
  author    = {Zheng, Jinliang and Li, Jianxiong and Wang, Zhihao and Liu, Dongxiu and Kang, Xirui and Feng, Yuchun and Zheng, Yinan and Zou, Jiayin and Chen, Yilun and Zeng, Jia and Wang, Tai and Zhang, Ya-Qin and Liu, Jingjing and Zhan, Xianyuan},
  booktitle = {International Conference on Learning Representations (ICLR)},
  year      = {2026},
  url       = {https://openreview.net/forum?id=kt51kZH4aG},
  note      = {arXiv:2510.10274}
}

@inproceedings{metaworld2019,
  title     = {Meta-World: A Benchmark and Evaluation for Multi-Task and Meta Reinforcement Learning},
  author    = {Yu, Tianhe and Quillen, Deirdre and He, Zhanpeng and Julian, Ryan and Hausman, Karol and Finn, Chelsea and Levine, Sergey},
  booktitle = {Proceedings of the Conference on Robot Learning},
  series    = {Proceedings of Machine Learning Research},
  volume    = {100},
  pages     = {1094--1100},
  publisher = {PMLR},
  year      = {2020},
  url       = {https://proceedings.mlr.press/v100/yu20a.html}
}

@article{xpolicylab2026,
  title   = {{XPolicyLab}: A Unified Standard and Open Ecosystem for Robot Policy Evaluation and Deployment},
  author  = {{XPolicyLab Community} and Chen, Tianxing and Chen, Yue and Nian, Tian and Cai, Zijian and Chen, Guangyu and Lin, Wenwei and Liang, Qiwei and Chen, Zanxin and Xiang, Peicheng and Su, Kailun and Li, Zixuan and Tang, Junyuan and Qin, Yan and Chen, Qiangyu and Zhu, Shaolong and Jiang, Tengyue and Wang, Yiqing and Li, Xiang and Zhang, Jiahao and Wan, Weijie and Chen, Baijun and Su, Honghao and Ye, Kehe and Liu, Shujia and Wang, Kaixuan and Liang, Haotian and Liu, Yunze and Li, Mingleyang and Wang, Yuran and Chen, Boyu and Bi, Hongzhe and Huang, Shuhe and Tan, Hengkai and Cai, Jisong and Mu, Yao and Guo, Jun and Wang, Xiaofeng and Zhu, Zheng and Ke, Weijie and Li, Hengtao and Tang, Yuhang and Li, Xiaofan and Yang, Ganlin and Tu, Zhangzheng and Yang, Shuai and Song, Wenxuan and Ding, Pengxiang and Zhang, Kaidong and Sun, Yu and Guo, Junliang and Zhang, Tong and Chen, Yixing and Cui, Rongxu and Zhang, Zongzheng and Ma, Haoxiang and Cai, Junhao and Zhang, Haoyu and Yang, Senqiao and Ye, Jinhui and Chen, Pengguang and Liu, Shu and Su, Xiu and Fang, Wenhan and Li, Wenhao and Cao, Yichao and Wang, Chengyao and Chen, Qiang and Luo, Ping and Ding, Wenbo},
  journal = {arXiv preprint arXiv:2608.09892},
  year    = {2026}
}

@article{robodojo2026,
  title   = {{RoboDojo}: A Unified Sim-and-Real Benchmark for Comprehensive Evaluation of Generalist Robot Manipulation Policies},
  author  = {Chen, Tianxing and Chen, Yue and Li, Zixuan and Tang, Junyuan and Su, Kailun and Lu, Haoran and Wan, Weijie and Chen, Baijun and Liu, Songling and Yan, Haowen and Su, Honghao and Dou, Zhiyang and Wang, Kaixuan and Zhang, Dandan and Liu, Yunze and Qin, Yan and Liang, Qiwei and Wu, Qiwei and Lin, Zijian and Lin, Wenwei and Wang, Yuran and He, Minghua and Wu, Tianshu and Wu, Ruihai and Zhou, Jingquan and Lei, Kai-Chong and Yu, Haibao and Ji, Yuanfeng and Jin, Weiyang and Lin, Guanyu and Li, Xiaofan and Xiong, Qi and Xu, Renjing and Li, Zhongyu and Chai, Wenhao and Xie, Enze and Wang, Ziwei and Mu, Yao and Dong, Hao and Matusik, Wojciech and Ding, Mingyu and Ding, Wenbo and Luo, Ping and Tomizuka, Masayoshi},
  journal = {arXiv preprint arXiv:2607.04434},
  year    = {2026}
}

@article{liberopro2025,
  title   = {{LIBERO-PRO}: Towards Robust and Fair Evaluation of Vision-Language-Action Models Beyond Memorization},
  author  = {Zhou, Xueyang and Xu, Yangming and Tie, Guiyao and Chen, Yongchao and Zhang, Guowen and Chu, Duanfeng and Zhou, Pan and Sun, Lichao},
  journal = {arXiv preprint arXiv:2510.03827},
  year    = {2025}
}

@article{rlbench2020,
  title   = {{RLBench}: The Robot Learning Benchmark \& Learning Environment},
  author  = {James, Stephen and Ma, Zicong and Arrojo, David Rovick and Davison, Andrew J.},
  journal = {IEEE Robotics and Automation Letters},
  volume  = {5},
  number  = {2},
  pages   = {3019--3026},
  year    = {2020}
}

@inproceedings{peract2022,
  title     = {{Perceiver-Actor}: A Multi-Task Transformer for Robotic Manipulation},
  author    = {Shridhar, Mohit and Manuelli, Lucas and Fox, Dieter},
  booktitle = {Proceedings of The 6th Conference on Robot Learning},
  series    = {Proceedings of Machine Learning Research},
  volume    = {205},
  pages     = {785--799},
  publisher = {PMLR},
  year      = {2023},
  url       = {https://proceedings.mlr.press/v205/shridhar23a.html}
}

@article{graham1969bounds,
  author  = {Graham, Ronald L.},
  title   = {Bounds on Multiprocessing Timing Anomalies},
  journal = {SIAM Journal on Applied Mathematics},
  volume  = {17},
  number  = {2},
  pages   = {416--429},
  year    = {1969},
  doi     = {10.1137/0117039}
}

@inproceedings{libero2023,
  title     = {{LIBERO}: Benchmarking Knowledge Transfer for Lifelong Robot Learning},
  author    = {Liu, Bo and Zhu, Yifeng and Gao, Chongkai and Feng, Yihao and Liu, Qiang and Zhu, Yuke and Stone, Peter},
  booktitle = {Advances in Neural Information Processing Systems},
  volume    = {36},
  pages     = {44776--44791},
  year      = {2023},
  doi       = {10.52202/075280-1939}
}

@article{calvin2022,
  title   = {{CALVIN}: A Benchmark for Language-Conditioned Policy Learning for Long-Horizon Robot Manipulation Tasks},
  author  = {Mees, Oier and Hermann, Lukas and Rosete-Beas, Erick and Burgard, Wolfram},
  journal = {IEEE Robotics and Automation Letters},
  volume  = {7},
  number  = {3},
  pages   = {7327--7334},
  year    = {2022},
  url     = {https://arxiv.org/abs/2112.03227}
}

@inproceedings{robotwin2025,
  title     = {{RoboTwin}: Dual-Arm Robot Benchmark with Generative Digital Twins},
  author    = {Mu, Yao and Chen, Tianxing and Chen, Zanxin and Peng, Shijia and Lan, Zhiqian and Gao, Zeyu and Liang, Zhixuan and Yu, Qiaojun and Zou, Yude and Xu, Mingkun and Lin, Lunkai and Xie, Zhiqiang and Ding, Mingyu and Luo, Ping},
  booktitle = {Proceedings of the IEEE/CVF Conference on Computer Vision and Pattern Recognition (CVPR)},
  pages     = {27649--27660},
  year      = {2025},
  doi       = {10.1109/CVPR52734.2025.02575},
  url       = {https://openaccess.thecvf.com/content/CVPR2025/html/Mu_RoboTwin_Dual-Arm_Robot_Benchmark_with_Generative_Digital_Twins_CVPR_2025_paper.html}
}

@inproceedings{robocasa2024,
  title     = {{RoboCasa}: Large-Scale Simulation of Household Tasks for Generalist Robots},
  author    = {Nasiriany, Soroush and Maddukuri, Abhiram and Zhang, Lance and Parikh, Adeet and Lo, Aaron and Joshi, Abhishek and Mandlekar, Ajay and Zhu, Yuke},
  booktitle = {Robotics: Science and Systems XX},
  year      = {2024},
  doi       = {10.15607/RSS.2024.XX.050}
}

@misc{starvla2026,
  title         = {{StarVLA}: A Lego-like Codebase for Vision-Language-Action Model Developing},
  author        = {{StarVLA Community}},
  year          = {2026},
  eprint        = {2604.05014},
  archivePrefix = {arXiv},
  primaryClass  = {cs.RO},
  url           = {https://arxiv.org/abs/2604.05014}
}

@inproceedings{henderson2018,
  title     = {Deep Reinforcement Learning That Matters},
  author    = {Henderson, Peter and Islam, Riashat and Bachman, Philip and Pineau, Joelle and Precup, Doina and Meger, David},
  booktitle = {Proceedings of the AAAI Conference on Artificial Intelligence},
  volume    = {32},
  year      = {2018},
  doi       = {10.1609/aaai.v32i1.11694}
}

@inproceedings{liberoplus2025,
    title={{LIBERO-Plus}: A Progressive Robustness Benchmark for Visual-Language-Action Models},
    author={Senyu Fei and Siyin Wang and Junhao Shi and Zihao Dai and Jikun Cai and Pengfang Qian and Li Ji and Xinzhe He and Shiduo Zhang and Zhaoye Fei and Jinlan Fu and Jingjing Gong and Xipeng Qiu},
    booktitle={Proceedings of the IEEE/CVF Conference on Computer Vision and Pattern Recognition (CVPR)},
    pages={38574--38583},
    year={2026},
    url={https://openaccess.thecvf.com/content/CVPR2026/html/Fei_LIBERO-Plus_A_Progressive_Robustness_Benchmark_for_Visual-Language-Action_Models_CVPR_2026_paper.html},
}

@misc{gr00t2025,
      title={{GR00T N1}: An Open Foundation Model for Generalist Humanoid Robots},
      author={NVIDIA and Johan Bjorck and Fernando Casta{\~n}eda and Nikita Cherniadev and Xingye Da and Runyu Ding and Linxi "Jim" Fan and Yu Fang and Dieter Fox and Fengyuan Hu and Spencer Huang and Joel Jang and Zhenyu Jiang and Jan Kautz and Kaushil Kundalia and Lawrence Lao and Zhiqi Li and Zongyu Lin and Kevin Lin and Guilin Liu and Edith Llontop and Loic Magne and Ajay Mandlekar and Avnish Narayan and Soroush Nasiriany and Scott Reed and You Liang Tan and Guanzhi Wang and Zu Wang and Jing Wang and Qi Wang and Jiannan Xiang and Yuqi Xie and Yinzhen Xu and Zhenjia Xu and Seonghyeon Ye and Zhiding Yu and Ao Zhang and Hao Zhang and Yizhou Zhao and Ruijie Zheng and Yuke Zhu},
      year={2025},
      eprint={2503.14734},
      archivePrefix={arXiv},
      primaryClass={cs.RO},
      url={https://arxiv.org/abs/2503.14734},
}

@inproceedings{robotwin22025,
  title = {{RoboTwin 2.0: A Scalable Data Generator and Benchmark with Strong Domain Randomization for Robust Bimanual Robotic Manipulation}},
  author = {Chen, Tianxing and Chen, Zanxin and Chen, Baijun and Cai, Zijian and Liu, Yibin and Li, Zixuan and Liang, Qiwei and Lin, Xianliang and Ge, Yiheng and Gu, Zhenyu and Deng, Weiliang and Guo, Yubin and Nian, Tian and Xie, Xuanbing and Chen, Qiangyu and Su, Kailun and Xu, Tianling and Liu, Guodong and Hu, Mengkang and Gao, Huan-Ang and Wang, Kaixuan and Liang, Zhixuan and Qin, Yusen and Yang, Xiaokang and Luo, Ping and Mu, Yao},
  year = {2026},
  booktitle = {Proceedings of the 43rd International Conference on Machine Learning},
  volume = {306},
  pages = {13673--13699},
  series = {Proceedings of Machine Learning Research},
  publisher = {PMLR},
  url = {https://proceedings.mlr.press/v306/chen26j.html},
}

@inproceedings{mujoco2012,
  title = {{MuJoCo: A physics engine for model-based control}},
  author = {Todorov, Emanuel and Erez, Tom and Tassa, Yuval},
  year = {2012},
  doi = {10.1109/IROS.2012.6386109},
  url = {https://doi.org/10.1109/IROS.2012.6386109},
  booktitle = {2012 IEEE/RSJ International Conference on Intelligent Robots and Systems},
  pages = {5026--5033},
}

@misc{robosuite2020,
  title = {{robosuite: A Modular Simulation Framework and Benchmark for Robot Learning}},
  author = {Zhu, Yuke and Wong, Josiah and Mandlekar, Ajay and Martín-Martín, Roberto and Joshi, Abhishek and Lin, Kevin and Maddukuri, Abhiram and Nasiriany, Soroush and Zhu, Yifeng},
  year = {2020},
  eprint = {2009.12293},
  archivePrefix = {arXiv},
  url = {https://arxiv.org/abs/2009.12293},
  note = {Revised 2025-01-18},
}

@inproceedings{sapien2020,
  title = {{SAPIEN: A SimulAted Part-based Interactive ENvironment}},
  author = {Xiang, Fanbo and Qin, Yuzhe and Mo, Kaichun and Xia, Yikuan and Zhu, Hao and Liu, Fangchen and Liu, Minghua and Jiang, Hanxiao and Yuan, Yifu and Wang, He and Yi, Li and Chang, Angel X. and Guibas, Leonidas J. and Su, Hao},
  booktitle = {Proceedings of the IEEE/CVF Conference on Computer Vision and Pattern Recognition (CVPR)},
  pages = {11094--11104},
  year = {2020},
  doi = {10.1109/CVPR42600.2020.01111},
  eprint = {2003.08515},
  archivePrefix = {arXiv},
  url = {https://arxiv.org/abs/2003.08515},
}

@inproceedings{gymnasium2025,
  title = {{Gymnasium: A Standard Interface for Reinforcement Learning Environments}},
  author = {Towers, Mark and Kwiatkowski, Ariel and Balis, John U. and De Cola, Gianluca and Deleu, Tristan and Goulão, Manuel and Kallinteris, Andreas and Krimmel, Markus and KG, Arjun and Perez-Vicente, Rodrigo and Terry, Jordan and Pierré, Andrea and Schulhoff, Sander and Tai, Jun Jet and Tan, Hannah and Younis, Omar G.},
  year = {2025},
  booktitle = {Advances in Neural Information Processing Systems},
  volume = {38},
  pages = {163114--163129},
  doi = {10.52202/085713-4916},
  url = {https://proceedings.neurips.cc/paper_files/paper/2025/hash/d7ff1795e8527f6443371c3933bdb52b-Abstract-Datasets_and_Benchmarks_Track.html},
}

@inproceedings{robomimic2021,
  title = {{What Matters in Learning from Offline Human Demonstrations for Robot Manipulation}},
  author = {Mandlekar, Ajay and Xu, Danfei and Wong, Josiah and Nasiriany, Soroush and Wang, Chen and Kulkarni, Rohun and Fei-Fei, Li and Savarese, Silvio and Zhu, Yuke and Martín-Martín, Roberto},
  booktitle = {Proceedings of the 5th Conference on Robot Learning},
  volume = {164},
  pages = {1678--1690},
  series = {Proceedings of Machine Learning Research},
  publisher = {PMLR},
  year = {2022},
  url = {https://proceedings.mlr.press/v164/mandlekar22a.html},
}

@inproceedings{openvla2024,
  title = {{OpenVLA: An Open-Source Vision-Language-Action Model}},
  author = {Kim, Moo Jin and Pertsch, Karl and Karamcheti, Siddharth and Xiao, Ted and Balakrishna, Ashwin and Nair, Suraj and Rafailov, Rafael and Foster, Ethan P. and Sanketi, Pannag R. and Vuong, Quan and Kollar, Thomas and Burchfiel, Benjamin and Tedrake, Russ and Sadigh, Dorsa and Levine, Sergey and Liang, Percy and Finn, Chelsea},
  booktitle = {Proceedings of the 8th Conference on Robot Learning (CoRL)},
  series = {Proceedings of Machine Learning Research},
  volume = {270},
  pages = {2679--2713},
  publisher = {PMLR},
  year = {2025},
  eprint = {2406.09246},
  archivePrefix = {arXiv},
  url = {https://proceedings.mlr.press/v270/kim25c.html},
}

@inproceedings{oft2025,
  title = {{Fine-Tuning Vision-Language-Action Models: Optimizing Speed and Success}},
  author = {Kim, Moo Jin and Finn, Chelsea and Liang, Percy},
  booktitle = {Robotics: Science and Systems (RSS)},
  year = {2025},
  eprint = {2502.19645},
  archivePrefix = {arXiv},
  url = {https://arxiv.org/abs/2502.19645},
}

@inproceedings{octo2024,
  title = {{Octo: An Open-Source Generalist Robot Policy}},
  author = {Ghosh, Dibya and Walke, Homer Rich and Pertsch, Karl and Black, Kevin and Mees, Oier and Dasari, Sudeep and Hejna, Joey and Kreiman, Tobias and Xu, Charles and Luo, Jianlan and Tan, You Liang and Chen, Lawrence Yunliang and Vuong, Quan and Xiao, Ted and Sanketi, Pannag R. and Sadigh, Dorsa and Finn, Chelsea and Levine, Sergey},
  booktitle = {Robotics: Science and Systems (RSS)},
  year = {2024},
  doi = {10.15607/RSS.2024.XX.090},
  eprint = {2405.12213},
  archivePrefix = {arXiv},
  url = {https://www.roboticsproceedings.org/rss20/p090.html},
}

@inproceedings{pi02025,
  title = {{$\pi_0$: A Vision-Language-Action Flow Model for General Robot Control}},
  author = {Black, Kevin and Brown, Noah and Driess, Danny and Esmail, Adnan and Equi, Michael Robert and Finn, Chelsea and Fusai, Niccolo and Groom, Lachy and Hausman, Karol and Ichter, Brian and Jakubczak, Szymon and Jones, Tim and Ke, Liyiming and Levine, Sergey and Li-Bell, Adrian and Mothukuri, Mohith and Nair, Suraj and Pertsch, Karl and Shi, Lucy Xiaoyang and Smith, Laura and Tanner, James and Vuong, Quan and Walling, Anna and Wang, Haohuan and Zhilinsky, Ury},
  booktitle = {Robotics: Science and Systems (RSS)},
  year = {2025},
  doi = {10.15607/RSS.2025.XXI.010},
  url = {https://www.roboticsproceedings.org/rss21/p010.html},
}

@misc{qwen25vl2025,
  title = {{Qwen2.5-VL Technical Report}},
  author = {Bai, Shuai and Chen, Keqin and Liu, Xuejing and Wang, Jialin and Ge, Wenbin and Song, Sibo and Dang, Kai and Wang, Peng and Wang, Shijie and Tang, Jun and Zhong, Humen and Zhu, Yuanzhi and Yang, Mingkun and Li, Zhaohai and Wan, Jianqiang and Wang, Pengfei and Ding, Wei and Fu, Zheren and Xu, Yiheng and Ye, Jiabo and Zhang, Xi and Xie, Tianbao and Cheng, Zesen and Zhang, Hang and Yang, Zhibo and Xu, Haiyang and Lin, Junyang},
  year = {2025},
  eprint = {2502.13923},
  archivePrefix = {arXiv},
  url = {https://arxiv.org/abs/2502.13923},
}

@misc{qwen3vl2025,
  title = {{Qwen3-VL Technical Report}},
  author = {Bai, Shuai and Cai, Yuxuan and Chen, Ruizhe and Chen, Keqin and Chen, Xionghui and Cheng, Zesen and Deng, Lianghao and Ding, Wei and Gao, Chang and Ge, Chunjiang and Ge, Wenbin and Guo, Zhifang and Huang, Qidong and Huang, Jie and Huang, Fei and Hui, Binyuan and Jiang, Shutong and Li, Zhaohai and Li, Mingsheng and Li, Mei and Li, Kaixin and Lin, Zicheng and Lin, Junyang and Liu, Xuejing and Liu, Jiawei and Liu, Chenglong and Liu, Yang and Liu, Dayiheng and Liu, Shixuan and Lu, Dunjie and Luo, Ruilin and Lv, Chenxu and Men, Rui and Meng, Lingchen and Ren, Xuancheng and Ren, Xingzhang and Song, Sibo and Sun, Yuchong and Tang, Jun and Tu, Jianhong and Wan, Jianqiang and Wang, Peng and Wang, Pengfei and Wang, Qiuyue and Wang, Yuxuan and Xie, Tianbao and Xu, Yiheng and Xu, Haiyang and Xu, Jin and Yang, Zhibo and Yang, Mingkun and Yang, Jianxin and Yang, An and Yu, Bowen and Zhang, Fei and Zhang, Hang and Zhang, Xi and Zheng, Bo and Zhong, Humen and Zhou, Jingren and Zhou, Fan and Zhou, Jing and Zhu, Yuanzhi and Zhu, Ke},
  year = {2025},
  eprint = {2511.21631},
  archivePrefix = {arXiv},
  url = {https://arxiv.org/abs/2511.21631},
}

@inproceedings{rliable2021,
  title = {{Deep Reinforcement Learning at the Edge of the Statistical Precipice}},
  author = {Agarwal, Rishabh and Schwarzer, Max and Castro, Pablo Samuel and Courville, Aaron and Bellemare, Marc G.},
  booktitle = {Advances in Neural Information Processing Systems},
  volume = {34},
  pages = {29304--29320},
  year = {2021},
  eprint = {2108.13264},
  archivePrefix = {arXiv},
  url = {https://arxiv.org/abs/2108.13264},
}

@article{diffusionpolicy2025,
  title = {{Diffusion Policy: Visuomotor Policy Learning via Action Diffusion}},
  author = {Chi, Cheng and Xu, Zhenjia and Feng, Siyuan and Cousineau, Eric and Du, Yilun and Burchfiel, Benjamin and Tedrake, Russ and Song, Shuran},
  year = {2025},
  journal = {The International Journal of Robotics Research},
  volume = {44},
  number = {10--11},
  pages = {1684--1704},
  doi = {10.1177/02783649241273668},
  url = {https://journals.sagepub.com/doi/10.1177/02783649241273668},
}

@inproceedings{act2023,
  title = {{Learning Fine-Grained Bimanual Manipulation with Low-Cost Hardware}},
  author = {Zhao, Tony Z. and Kumar, Vikash and Levine, Sergey and Finn, Chelsea},
  booktitle = {Robotics: Science and Systems (RSS)},
  year = {2023},
  eprint = {2304.13705},
  archivePrefix = {arXiv},
  url = {https://arxiv.org/abs/2304.13705},
}

@inproceedings{isaacgym2021,
  title = {{Isaac Gym: High Performance GPU Based Physics Simulation For Robot Learning}},
  author = {Makoviychuk, Viktor and Wawrzyniak, Lukasz and Guo, Yunrong and Lu, Michelle and Storey, Kier and Macklin, Miles and Hoeller, David and Rudin, Nikita and Allshire, Arthur and Handa, Ankur and State, Gavriel},
  booktitle = {Proceedings of the Neural Information Processing Systems Track on Datasets and Benchmarks},
  volume = {1},
  year = {2021},
  eprint = {2108.10470},
  archivePrefix = {arXiv},
  url = {https://datasets-benchmarks-proceedings.neurips.cc/paper/2021/hash/28dd2c7955ce926456240b2ff0100bde-Abstract-round2.html},
}

@inproceedings{maniskill32024,
  title = {{Demonstrating GPU Parallelized Robot Simulation and Rendering for Generalizable Embodied AI with ManiSkill3}},
  author = {Tao, Stone and Xiang, Fanbo and Shukla, Arth and Qin, Yuzhe and Hinrichsen, Xander and Yuan, Xiaodi and Bao, Chen and Lin, Xinsong and Liu, Yulin and Chan, Tse-Kai and Gao, Yuan and Li, Xuanlin and Mu, Tongzhou and Xiao, Nan and Gurha, Arnav and Rajesh, Viswesh Nagaswamy and Choi, Yong Woo and Chen, Yen-Ru and Huang, Zhiao and Calandra, Roberto and Chen, Rui and Luo, Shan and Su, Hao},
  booktitle = {Proceedings of Robotics: Science and Systems},
  volume = {XXI},
  year = {2025},
  doi = {10.15607/RSS.2025.XXI.021},
  eprint = {2410.00425},
  archivePrefix = {arXiv},
  url = {https://roboticsproceedings.org/rss21/p021.html},
}

@inproceedings{envpool2022,
  title = {{EnvPool: A Highly Parallel Reinforcement Learning Environment Execution Engine}},
  author = {Weng, Jiayi and Lin, Min and Huang, Shengyi and Liu, Bo and Makoviichuk, Denys and Makoviychuk, Viktor and Liu, Zichen and Song, Yufan and Luo, Ting and Jiang, Yukun and Xu, Zhongwen and Yan, Shuicheng},
  booktitle = {Advances in Neural Information Processing Systems},
  volume = {35},
  pages = {22409--22421},
  year = {2022},
  doi = {10.52202/068431-1628},
  eprint = {2206.10558},
  archivePrefix = {arXiv},
  url = {https://arxiv.org/abs/2206.10558},
}

@inproceedings{impala2018,
  title = {{IMPALA: Scalable Distributed Deep-RL with Importance Weighted Actor-Learner Architectures}},
  author = {Espeholt, Lasse and Soyer, Hubert and Munos, Remi and Simonyan, Karen and Mnih, Volodymyr and Ward, Tom and Doron, Yotam and Firoiu, Vlad and Harley, Tim and Dunning, Iain and Legg, Shane and Kavukcuoglu, Koray},
  booktitle = {International Conference on Machine Learning},
  series = {Proceedings of Machine Learning Research},
  volume = {80},
  pages = {1407--1416},
  year = {2018},
  eprint = {1802.01561},
  archivePrefix = {arXiv},
  url = {https://arxiv.org/abs/1802.01561},
}

@inproceedings{seedrl2020,
  title = {{SEED RL: Scalable and Efficient Deep-RL with Accelerated Central Inference}},
  author = {Espeholt, Lasse and Marinier, Rapha{\"e}l and Stanczyk, Piotr and Wang, Ke and Michalski, Marcin},
  booktitle = {International Conference on Learning Representations},
  year = {2020},
  eprint = {1910.06591},
  archivePrefix = {arXiv},
  url = {https://arxiv.org/abs/1910.06591},
}

@inproceedings{samplefactory2020,
  title = {{Sample Factory: Egocentric 3D Control from Pixels at 100000 FPS with Asynchronous Reinforcement Learning}},
  author = {Petrenko, Aleksei and Huang, Zhehui and Kumar, Tushar and Sukhatme, Gaurav and Koltun, Vladlen},
  booktitle = {International Conference on Machine Learning},
  series = {Proceedings of Machine Learning Research},
  volume = {119},
  pages = {7652--7662},
  year = {2020},
  eprint = {2006.11751},
  archivePrefix = {arXiv},
  url = {https://arxiv.org/abs/2006.11751},
}

@inproceedings{simplerenv2025,
  title = {{Evaluating Real-World Robot Manipulation Policies in Simulation}},
  author = {Li, Xuanlin and Hsu, Kyle and Gu, Jiayuan and Mees, Oier and Pertsch, Karl and Walke, Homer Rich and Fu, Chuyuan and Lunawat, Ishikaa and Sieh, Isabel and Kirmani, Sean and Levine, Sergey and Wu, Jiajun and Finn, Chelsea and Su, Hao and Vuong, Quan and Xiao, Ted},
  year = {2025},
  booktitle = {Proceedings of The 8th Conference on Robot Learning},
  volume = {270},
  pages = {3705--3728},
  series = {Proceedings of Machine Learning Research},
  publisher = {PMLR},
  url = {https://proceedings.mlr.press/v270/li25c.html},
}

\appendix
\section{Policy and Benchmark Contracts}
\label{app:v2-contracts}
\subsection{Public Policy Lifecycle}
\begin{figure}[tbp]
\centering
\includegraphics[width=0.92\linewidth]{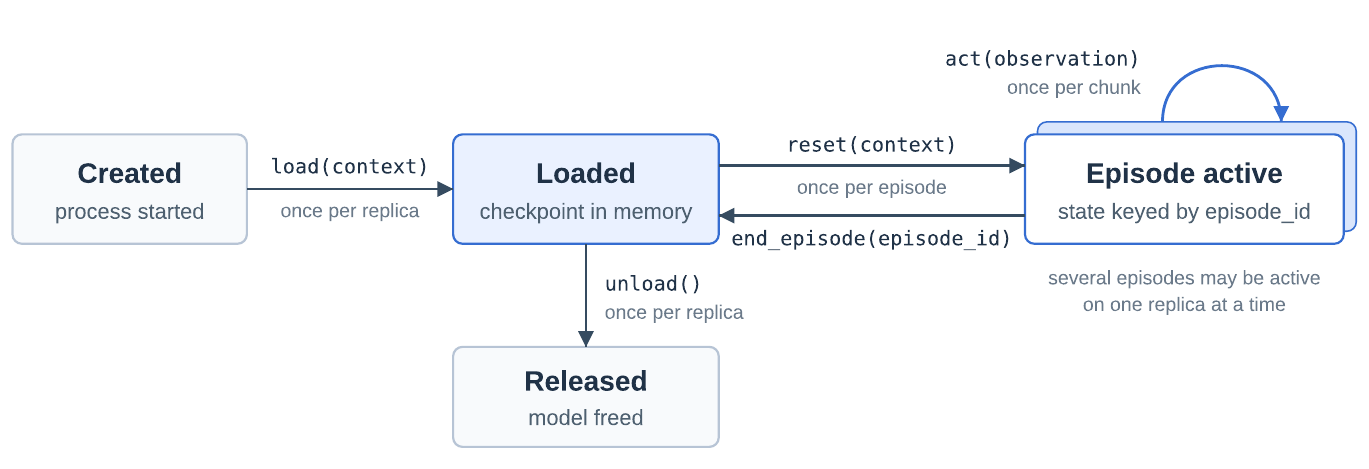}
\caption{Policy lifecycle within one replica. The checkpoint is loaded once; each episode adds and
later removes its own state, and several episodes may be active at a time.}
\label{fig:v2-lifecycle}
\end{figure}

Figure~\ref{fig:v2-lifecycle} shows the lifecycle of one replica, and
Table~\ref{tab:v2-lifecycle} lists the lifecycle methods. The Control Runtime invokes \code{load} once per
replica process. Several simulator workers interleave episodes through one replica, so stateful
Policies use the episode identifier as their state key.

\begin{table}[tbp]
\centering
\small
\caption{Policy lifecycle methods and ownership.}
\label{tab:v2-lifecycle}
\begin{tabularx}{\linewidth}{@{}>{\raggedright\arraybackslash}p{40mm} >{\raggedright\arraybackslash}p{36mm} X@{}}
\toprule
Method & Lifecycle transition & Policy behavior \\
\midrule
\code{load(context)} & Created $\rightarrow$ Loaded & Load one checkpoint replica and declare capabilities \\
\code{reset(context)} & Loaded $\rightarrow$ Episode active & Allocate state keyed by \code{episode\_id} \\
\code{act(observation)} & Episode active $\rightarrow$ Episode active & Produce one action or action chunk from canonical features \\
\code{act\_batch} & Episode active $\rightarrow$ Episode active & Serve an observation batch with native batching when available \\
\code{end\_episode(episode\_id)} & Episode active $\rightarrow$ Loaded & Release state for one episode \\
\code{unload()} & Loaded $\rightarrow$ Released & Release model and process-wide resources \\
\bottomrule
\end{tabularx}
\end{table}

\subsection{Policy Examples}
\label{app:v2-policy-examples}
What a Policy must add beyond \code{load} and \code{act} depends on whether it keeps state between
inference calls. Listing~\ref{lst:v2-stateless} shows a stateless Policy, condensed from the StarVLA
RoboTwin integration: each action chunk depends only on the current observation, so the class keeps
only replica-scope state, the model, and declares an observation stride equal to its chunk length,
which enables chunk-boundary observation (Section~\ref{sec:v2-runtime}).

\begin{lstlisting}[float=tbp,caption={The stateless StarVLA Policy, RoboTwin path (condensed).},label={lst:v2-stateless}]
from roboscope_kit import (ActionChunk, Observation, Policy,
                           PolicyCapabilities, SessionContext, run)

CAMERAS = ("images.head", "images.left_wrist", "images.right_wrist")

class StarVLARoboTwinPolicy(Policy):
    def load(self, context: SessionContext) -> PolicyCapabilities:
        self.model = load_checkpoint(context.checkpoint_path).to("cuda").eval()
        horizon = self.model.action_horizon             # 16 for this checkpoint
        return PolicyCapabilities(
            supported_benchmarks=("robotwin",),
            required_features=CAMERAS + ("language.instruction",),
            supported_action_chunk_sizes=(horizon,),
            observation_strides={"robotwin": horizon})  # render once per chunk

    def act(self, observation: Observation) -> ActionChunk:
        f = observation.features                        # canonical feature names
        example = preprocess([f[c] for c in CAMERAS], f["language.instruction"])
        normalized = self.model.predict_action(example)   # (16, 14), in [-1, 1]
        return ActionChunk(unnormalize(normalized).astype("float32"))

    def unload(self) -> None:
        self.model = None

if __name__ == "__main__":
    run(StarVLARoboTwinPolicy())
\end{lstlisting}

A stateful Policy carries information from one call to the next. The X-VLA Policy
\citep{xvla2026} is the stateful Policy integrated with \system{}: the
model takes as proprioceptive input the end-effector pose of its own previous action chunk, so each
episode carries a pose that every \code{act} reads and updates. Listing~\ref{lst:v2-stateful}
shows its LIBERO path. One replica
serves several simulator workers, and a replica handles one call at a time, so calls from
different episodes interleave on the same object. \rev{Per-episode state therefore cannot live in instance attributes. Instead, the Policy keys it by \code{episode\_id}, discards stale state in \code{reset}, initializes it from the observed state on the episode's first \code{act}, because \code{reset} carries no observation, and releases it in \code{end\_episode}.} Because the replica serializes calls,
the Policy needs no locking. The X-VLA Policy declares a stride of 1 on every benchmark.

\begin{lstlisting}[float=tbp,caption={The stateful X-VLA Policy, LIBERO path (condensed).},label={lst:v2-stateful}]
import numpy as np
from roboscope_kit import (ActionChunk, EpisodeContext, Observation,
                           Policy, PolicyCapabilities, SessionContext)

class XVLAPolicy(Policy):
    def __init__(self) -> None:
        self.proprio: dict[str, np.ndarray] = {}        # episode_id -> state

    def load(self, context: SessionContext) -> PolicyCapabilities:
        self.model = load_xvla(context.checkpoint_path).to("cuda").eval()
        return PolicyCapabilities(
            supported_benchmarks=("libero",),          # other benchmarks omitted
            required_features=("images.primary", "images.wrist",
                               "state.proprioception", "language.instruction"),
            supported_action_chunk_sizes=(self.model.num_actions,))

    def reset(self, context: EpisodeContext) -> None:
        self.proprio.pop(context.episode_id, None)     # no stale state

    def act(self, observation: Observation) -> ActionChunk:
        f, ep = observation.features, observation.episode_id
        if ep not in self.proprio:                     # first call of episode
            self.proprio[ep] = proprio_from_state(f["state.proprioception"])
        raw = self.model.generate(f["images.primary"], f["images.wrist"],
                                  f["language.instruction"], self.proprio[ep])
        self.proprio[ep][:9] = raw[-1, :9]               # feed back last pose
        return ActionChunk(to_libero_actions(raw))      # (30, 7) float32

    def end_episode(self, episode_id: str) -> None:
        self.proprio.pop(episode_id, None)              # release episode state
\end{lstlisting}

\subsection{Evaluation Data Types}
\begin{description}[itemsep=0.15em,topsep=0.3em,font=\normalfont\ttfamily]
  \item[FeatureSpec] canonical name, image/state/language kind, required flag, description, dtype, and optional shape.
  \item[ActionSpec] canonical action name, description, optional dimension, and dtype.
  \item[BenchmarkSpec] benchmark, suite, protocol version, ordered features, and action spec.
  \item[SessionContext] session identifier, checkpoint path, benchmark spec, and metadata.
  \item[EpisodeContext] episode identifier, task identifier, and benchmark metadata.
  \item[Observation] episode identifier, step index, named features, and optional timestamp.
  \item[ActionChunk] one action vector or a two-dimensional action sequence.
  \item[PolicyCapabilities] supported benchmarks and suites, required features, maximum batch,
  supported chunk lengths, action representations, and observation strides.
\end{description}
The Control Runtime validates that declared action representations and strides refer to supported
benchmarks and resolves one execution contract per benchmark, which it records with the run.

\subsection{Benchmark Contracts}
Table~\ref{tab:v2-contracts} summarizes the observation and action contracts.

\begin{table}[tbp]
\centering
\small
\caption{Canonical features and actions.}
\label{tab:v2-contracts}
\begin{tabularx}{\linewidth}{@{}>{\raggedright\arraybackslash}p{20mm} >{\raggedright\arraybackslash}p{30mm} X >{\raggedright\arraybackslash}p{40mm}@{}}
\toprule
Benchmark & Protocol & Required canonical features & Action contract \\
\midrule
LIBERO & \code{libero-v1} & \code{images.primary} $256\times256\times3$ uint8; \code{images.wrist}; 16-D float32 proprioception; instruction & 7-D float32 end-effector delta plus gripper \\
LIBERO-Plus & \code{libero-plus-v1} & Same named contract as LIBERO & 7-D float32 end-effector delta plus gripper \\
RoboTwin & \code{robotwin2-v1} & Head, left-wrist, and right-wrist RGB; dual-arm joint state; instruction & 14-D float32 absolute dual-arm joint and gripper target \\
RoboCasa & \code{robocasa-v1} & Primary RGB; optional wrist RGB; GR-1 state; instruction & 29-D float32 arms, hands, and waist target \\
CALVIN & \code{calvin-v1} & Static RGB; gripper RGB; proprioception; instruction & 7-D float32 relative end-effector and gripper action \\
MetaWorld & \code{metaworld-v1} & \code{images.primary} $480\times480\times3$; 4-D end-effector position and gripper aperture; instruction & 4-D float32 end-effector delta and gripper command \\
\rev{LIBERO-Pro} & \rev{\code{libero-pro-v1}} & \rev{Same named contract as LIBERO} & \rev{7-D float32 end-effector delta plus gripper} \\
\rev{RoboDojo} & \rev{\code{robodojo-v1}} & \rev{Head, left-wrist, and right-wrist RGB; 14-D dual-arm joint state; both end-effector poses; instruction} & \rev{14-D float32 absolute dual-arm joint and gripper target} \\
\rev{RLBench} & \rev{\code{rlbench-peract-v2}} & \rev{Front, two shoulder, and wrist RGB $128\times128\times3$ with point clouds; joint and gripper state; instruction} & \rev{End-effector pose reached by motion planning, and a discrete gripper command} \\
\rev{robomimic} & \rev{\code{robomimic-v1}} & \rev{Low-dimensional robosuite observations as float32 vectors, without images} & \rev{7-D float32 end-effector pose delta plus gripper for each arm} \\
\bottomrule
\end{tabularx}
\end{table}

\subsection{Work Queue State Machine}
A work item carries a unique ordinal and identifier, a grouping hint, a benchmark payload, and the
expected episode-record count. After completing an item, a worker first claims a pending item with
the same grouping hint. LIBERO groups items by task, so a worker keeps its task environment instead of
rebuilding it; each RoboTwin item has its own group, so an idle worker takes the next window of any
task. Queue directories encode state: \code{pending} holds available items
written by the Control Runtime; \code{running} holds atomically claimed items named with worker ownership;
\code{partial} holds append-only records for a running multi-episode item; and \code{completed} holds
idempotent item results used by aggregation. Completion validates the item and group identifiers and
the expected record count. A repeated completion with identical content is accepted; differing content
raises an integrity error. Worker recovery moves each unfinished claim back to \code{pending} and clears
its partial state.

\subsection{Recorded Evaluation Artifacts}
Episode records contain the episode key, success, steps, elapsed time, and benchmark metadata.
Metrics record inference batches, queue wait, prediction time, transport time, errors, and per-replica
summaries. GPU samples give utilization and memory per device; CPU samples give consumed cores and
cgroup throttling ratios. Run-level artifacts preserve software revisions, checkpoint and manifest
hashes, the selected $K/g$, the probe history, queue state, results, and analysis outputs.

\begin{figure}[tbp]
\centering
\begin{tikzpicture}[
  box/.style={draw=evoblue, rounded corners=2pt, fill=evolight, align=center,
    minimum height=15mm, text width=24mm, font=\small},
  wide/.style={box, text width=105mm, minimum height=11mm},
  flow/.style={-{Stealth[length=2mm]}, thick, draw=evoblue}
]
\node[wide] (manifest) at (0,0) {Frozen workload and build identity\\
  deterministic partition: instance~$i$ runs work item~$j$ when \mbox{$j \bmod N \mathrel{=} i$}};
\node[box] (a) at (-4.0,-2.0) {Instance $0$\\local queue\\$K_0$ workers\\$g_0$ replicas};
\node[box] (b) at (-1.1,-2.0) {Instance $1$\\local queue\\$K_1$ workers\\$g_1$ replicas};
\node[font=\small] (dots) at (1.35,-2.0) {$\cdots$};
\node[box] (c) at (3.95,-2.0) {Instance $N{-}1$\\local queue\\$K_{N-1}$ workers\\$g_{N-1}$ replicas};
\node[wide] (merge) at (0,-4.0) {Validate the shard set and aggregate benchmark results\\
  counts and official scores; retain each shard's provenance};
\foreach \n in {a, b, c} {
  \draw[flow] (manifest.south) -- ++(0,-0.25) -| (\n.north);
  \draw[flow] (\n.south) -- ++(0,-0.25) -| (merge.north);
}
\end{tikzpicture}
\caption{Two levels of parallel evaluation. Each instance runs a fixed subset of work items with its
own queue, simulator workers, and Policy replicas; the shard results are validated and merged into
one benchmark result.}
\label{fig:v2-multi-instance}
\end{figure}
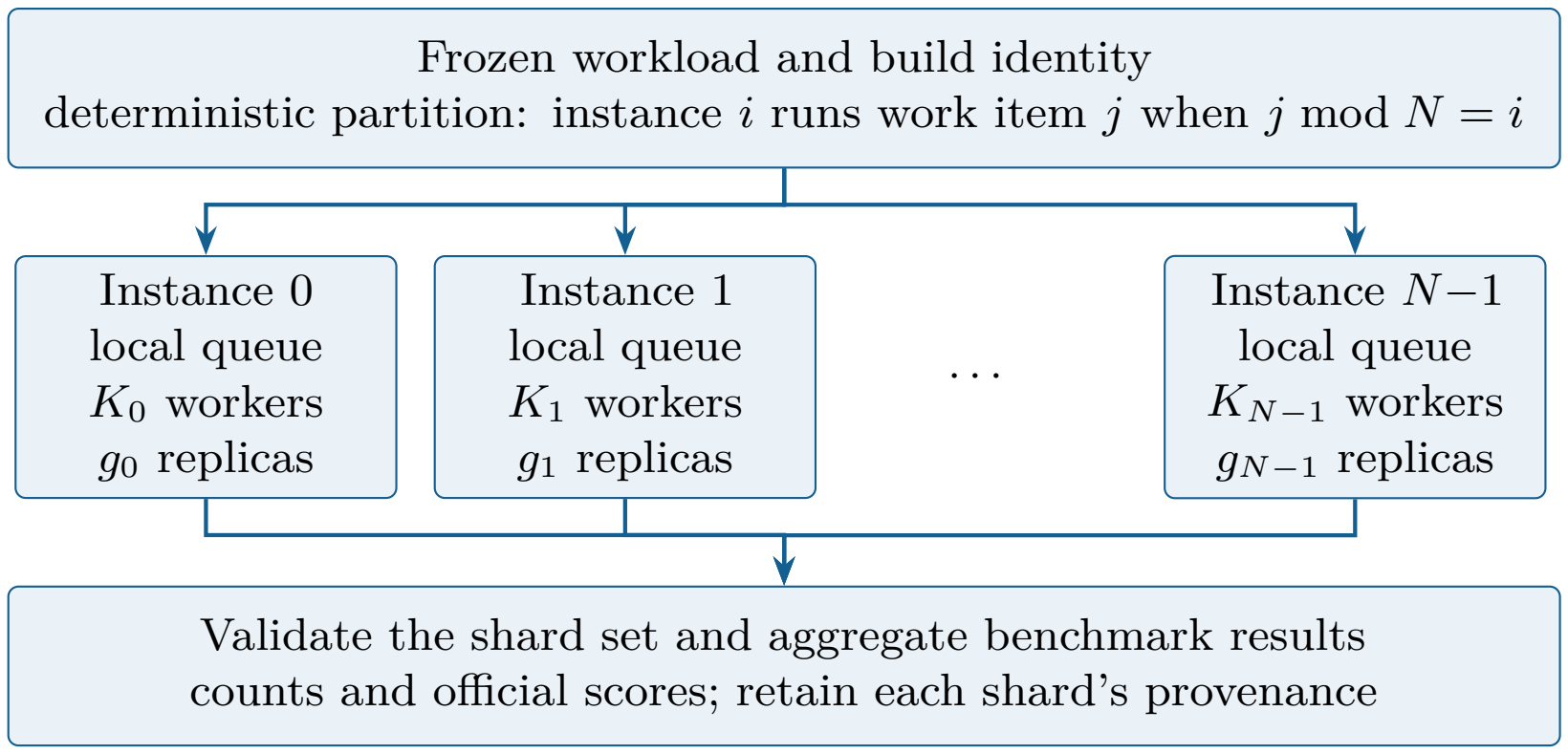

\section{Experimental Details}
\label{app:v2-setup}
\subsection{Workloads and Checkpoints}
LIBERO-Long uses task IDs 0--9, initial states 0--49, seed 7, ten settling steps, and EGL rendering,
with the published 30K-step checkpoint \nolinkurl{StarVLA/Qwen2.5-VL-OFT-LIBERO-4in1}. RoboTwin Easy uses
RoboTwin 2.0 \code{demo\_clean} with 50 tasks and episodes 0--9 per task, a validated seed cache, and
\nolinkurl{StarVLA/Qwen3-VL-OFT-Robotwin2}. The RoboTwin checkpoint carries
\nolinkurl{future_action_window_size=15}, which StarVLA schema 0.21 maps to \nolinkurl{action_horizon=16}.

\subsection{StarVLA Revisions}
\label{app:v2-revisions}
LIBERO-Long runs follow one of two StarVLA revisions with the same checkpoint and manifest. The
September 2026 runs follow \code{7e4a473} (30 June 2026), the upstream head at the time: the official
reference run and the fixed runs of Table~\ref{tab:v2-fixed-runs}. The October 2026 runs follow \code{eaa51c4} (24 February 2026),
the revision vla-eval v0.7.0 pins: the cross-system comparison, the worker and replica sweep of
Table~\ref{tab:v2-h1-scaling}, and the profiles of Section~\ref{sec:v2-time-profile}. The official
evaluation and vla-eval run the code of the batch's revision. \system{} builds the model from StarVLA
\code{7e4a473} in both batches, and its Policy applies the LIBERO client steps of the batch's
revision. The two LIBERO clients differ in three steps: image resizing (PIL
bilinear versus OpenCV area interpolation), action unnormalization (the server's training pipeline
versus min/max in the client), and the point at which the gripper command is binarized (after versus
before unnormalization). A fixed
$K1/g1$ \system{} run matches the batch's official evaluation on all 500 success outcomes: 439/500 under
\code{7e4a473} and 451/500 under \code{eaa51c4}.

\subsection{Baseline Configurations and Tuning}
The official StarVLA evaluation runs its policy server and LIBERO evaluator from StarVLA commit
\code{eaa51c4}, with one environment and one model. vla-eval runs release v0.7.0, which pins the same
commit behind its model server, and RLinf runs its main branch of 22 October 2026 with two local changes that align its
action unnormalization and gripper convention with the StarVLA checkpoint. All systems share the LIBERO-Long manifest, the checkpoint file, a chunk
length of 8, and the timing boundary of Section~\ref{sec:v2-methodology}.

RLinf was tuned once offline (Table~\ref{tab:v2-baseline-tuning}). Each probe evaluates 30
episodes, initial states 0--2 of every task, within a 30-minute search budget. A probe is eligible
if it covers all 30 episodes and has at least as many successes as the starting configuration;
RLinf keeps the eligible probe with the shortest 500-episode time projected from its probe epochs.
The starting configuration and the 20-environment configuration both scored 30, and the latter was
selected \rev{after 1,704.3\,s of tuning}.

\paragraph{Sampled actions in RLinf's StarVLA evaluation.}
\label{app:v2-rlinf-sampling}
RLinf's evaluation rollout samples StarVLA's actions instead of taking their mean. Its rollout worker
forwards the evaluation mode only to an allow-list of model types that excludes StarVLA, so the StarVLA
policy runs in its default training mode, which always enables sampling regardless of
\code{temperature\_eval}. The OFT action head then draws each normalized action from a Gaussian around the
predicted mean with standard deviation $\exp(-2.5)\approx 0.082$, on actions normalized to $[-1, 1]$. The
LIBERO-Long RLinf launches in this report sample in this way. Their sampled actions are
consistent with the lower agreement between repeated RLinf launches (86.2--88.4\%) than between repeated
launches of RoboScope Kit (Search) (97.8--99.0\%). No configuration value disables the sampling, and the evaluation
mode reaches every other model type in the allow-list, so this is likely a bug in RLinf's StarVLA
integration; it is present in RLinf's main branch as of 30 October 2026. Deterministic evaluation needs
RLinf to forward the evaluation mode to the StarVLA policy, a one-line change to the rollout worker. We
leave the LIBERO-Long launches as run, and the RoboTwin comparison applies this change so that RLinf
evaluates the mean action, as the other systems do (Appendix~\ref{app:v2-robotwin-baselines}).

vla-eval was tuned over its shard count, the number of model servers on the GPU, each server's
maximum batch size, and its batch wait (Table~\ref{tab:v2-vla-tuning}). With $g$ servers, shard $s$
connects to server $s \bmod g$. Probes evaluate 100 episodes, initial states 0--9 of every task, and
are ranked by steady-state throughput: policy steps per second from 30\,s after the shards start
until the first shard finishes. The first stage crosses 8 to 20 shards with one and two servers,
each batching up to the number of shards it serves with a 5\,ms wait. Throughput rises from
86.0\,steps/s at 8 shards to 133.0--135.8\,steps/s at 16 and 20, where 11.3 to 11.8 of the 12
cores are busy; a second server changes throughput by at most 2.6\%. The second stage varies the shard count, wait,
maximum batch, and server count around that point, and every probe falls within
131.7--136.8\,steps/s. The third stage runs all 500 episodes at three configurations of that range
and keeps the shortest evaluation time: 24 shards on one server, with batches of up to 24 and a
5\,ms wait, at 1,017.7\,s. Together with one earlier 500-episode run at 10 shards and two servers,
the 23 tuning runs took 9,780.1\,s.

\begin{table}[tbp]
\centering
\small
\caption{Offline tuning of the LIBERO-Long baselines.}
\label{tab:v2-baseline-tuning}
\begin{tabularx}{\linewidth}{@{}l Y P{36mm} r@{}}
\toprule
System & Search space & Selected & Runs \\
\midrule
vla-eval & Shards $\{8, \ldots, 24\}$, servers $\{1, 2, 3\}$, maximum batch $\{4, \ldots, 24\}$, wait
$\{0, 5, 20\}$\,ms; 100-episode probes, then 500-episode runs (Table~\ref{tab:v2-vla-tuning}) &
24 shards, 1 server, batch 24, wait 5\,ms & 23 \\
RLinf & Start: 10 environments, 1 pipeline stage; environments $\{4, 8, 12, 16, 20\}$; then 2 and 4
pipeline stages at 20 environments; 30-episode probes &
20 environments, 1 pipeline stage, 1 rollout worker (30/30) & 8 \\
\bottomrule
\end{tabularx}
\end{table}

\begin{table}[tbp]
\centering
\footnotesize
\setlength{\tabcolsep}{4pt}
\renewcommand{\arraystretch}{1.08}
\caption{vla-eval tuning runs on LIBERO-Long, in execution order. Stage 0 is the earlier 500-episode run.}
\label{tab:v2-vla-tuning}
\begin{tabular}{@{}c r r r r r r r r r@{}}
\toprule
Stage & Shards & Servers & Batch & Wait (ms) & Episodes & Successes & Eval.\ time (s) & Steady steps/s & Busy cores \\
\midrule
0 & 10 & 2 & 10 & 5 & 500 & 451 & 1,215.9 & -- & -- \\
1 & 10 & 1 & 10 & 5 & 100 & 90 & 317.1 & 107.9 & 9.4 \\
1 & 10 & 2 & 5 & 5 & 100 & 91 & 278.5 & 110.4 & 9.8 \\
1 & 8 & 1 & 8 & 5 & 100 & 88 & 355.2 & 86.0 & 7.7 \\
1 & 8 & 2 & 4 & 5 & 100 & 87 & 356.6 & 88.2 & 8.0 \\
1 & 12 & 1 & 12 & 5 & 100 & 88 & 318.9 & 123.6 & 10.7 \\
1 & 12 & 2 & 6 & 5 & 100 & 88 & 295.5 & 126.7 & 11.2 \\
1 & 16 & 1 & 16 & 5 & 100 & 90 & 235.6 & 133.6 & 11.3 \\
1 & 16 & 2 & 8 & 5 & 100 & 89 & 255.2 & 133.0 & 11.7 \\
1 & 20 & 1 & 20 & 5 & 100 & 89 & 233.7 & 135.8 & 11.5 \\
1 & 20 & 2 & 10 & 5 & 100 & 92 & 226.4 & 133.4 & 11.8 \\
2 & 24 & 1 & 24 & 5 & 100 & 89 & 245.8 & 136.8 & 11.6 \\
2 & 24 & 2 & 12 & 5 & 100 & 85 & 259.5 & 133.9 & 11.9 \\
2 & 14 & 1 & 14 & 5 & 100 & 90 & 266.7 & 131.7 & 11.1 \\
2 & 16 & 1 & 16 & 0 & 100 & 89 & 259.6 & 133.8 & 11.3 \\
2 & 16 & 1 & 16 & 20 & 100 & 89 & 250.8 & 133.6 & 11.3 \\
2 & 20 & 1 & 20 & 0 & 100 & 90 & 229.9 & 135.8 & 11.5 \\
2 & 20 & 1 & 20 & 20 & 100 & 92 & 226.6 & 134.4 & 11.5 \\
2 & 20 & 1 & 8 & 5 & 100 & 92 & 227.3 & 134.1 & 11.5 \\
2 & 20 & 3 & 7 & 5 & 100 & 90 & 227.4 & 132.5 & 11.9 \\
3 & 20 & 1 & 20 & 5 & 500 & 445 & 1,026.7 & 142.5 & 11.5 \\
3 & 24 & 1 & 24 & 5 & 500 & 450 & \textbf{1,017.7} & 141.8 & 11.6 \\
3 & 20 & 2 & 10 & 5 & 500 & 444 & 1,038.9 & 141.6 & 11.8 \\
\bottomrule
\end{tabular}
\par\smallskip
\raggedright All runs use the LIBERO-Long checkpoint, a chunk length of 8, and 12 pinned cores. Steady steps/s and busy cores are averages from 30\,s after the shards start until the first shard finishes. Batch is each server's maximum batch size.\par
\end{table}

\subsection{RoboTwin Cross-System Comparison}
\label{app:v2-robotwin-baselines}
All four systems load the \nolinkurl{StarVLA/Qwen3-VL-OFT-Robotwin2} checkpoint with StarVLA \code{7e4a473} and
evaluate the 500 episodes of RoboTwin \code{014728c} \code{demo\_clean}, the first ten seeds of each of the 50
tasks from the seed cache of Appendix~\ref{app:v2-setup}, with native per-task step limits. Every system takes
the instruction from the frozen manifest by \system{}'s rule: seed Python's generator with the episode seed,
generate the task's descriptions from the recorded episode information, and draw one unseen description. All
run on one RTX 4090 with a 20-core quota and no core pinning, one system at a time. Evaluation time is \system{}'s formal
evaluation pass, RLinf's rollout span, vla-eval's evaluation wall time, and, for the official evaluation, the sum of
its 500 episode times, each including the expert check; this leaves out the restart of its evaluator between tasks.
End-to-end time is the wall time of the whole launch, from model loading to teardown; for \system{}, policy
loading, runtime preparation, and the formal evaluation pass, excluding its 593\,s of autotuning, which
stopped at $K16$ on a worker OOM error. Throughput divides policy steps, 211,058 for all systems but
RLinf's 211,278, by evaluation time. 

\paragraph{Official StarVLA evaluation.}
The StarVLA RoboTwin evaluator runs with its default of one job per GPU. A patch forwards the checkpoint path, as
the StarVLA README does, takes the seeds and instructions from the frozen manifest, and writes one record per
episode; the official expert check still runs and is recorded.

\paragraph{RLinf.}
RLinf has no StarVLA path for RoboTwin, so an adapter of 14 recorded changes connects them. It loads the
checkpoint with \system{}'s loader, applies StarVLA's RoboTwin action unnormalization and image preprocessing, and
forwards the evaluation mode to the StarVLA policy, so that RLinf evaluates the mean action
(Appendix~\ref{app:v2-rlinf-sampling}). Its environments claim episodes from the frozen manifest in one launch,
execute each 16-action chunk with the official \code{take\_action} control loop, and render once per chunk; two
scene-setup differences in RLinf's RoboTwin fork are aligned with native RoboTwin. Inference is batched over
all environments, and each environment builds a new SAPIEN engine for every episode. Tuning selected 8
environments in 8 environment processes with one model replica.

\paragraph{vla-eval.}
vla-eval v0.7.0 runs with its static shard split, synchronous episode runner, and a new RoboTwin environment
with its own engine and renderer per episode. Eleven recorded changes run it outside its container on
Python~3.8, give it the frozen episodes and instructions, the native step limits, and StarVLA's RoboTwin action
post-processing, and load the checkpoint with \system{}'s loader. \rev{Stock vla-eval keeps each finished
episode's renderer and cameras, so its GPU memory grows with every episode; the formal launch releases them
before each new environment, which keeps each shard at 3.3\,GiB with the same outcomes.} Tuning selected 9 shards on
one model server with batch size~1, since batched inference changed outcomes.

\paragraph{Object-variant selection on the host.}
Several RoboTwin tasks pick an object model variant from an unsorted directory listing, so the variant built
for a seed depends on the order in which the host's file system lists the asset copy. On this host, 18 of the
500 seeds build a different variant from the one the seed cache recorded, and the manifest instruction can then
describe the recorded variant rather than the one in the scene. All four systems read the same asset directory
order, so they evaluate the same scenes.

\section{Evaluation-Time Profile}
\label{app:v2-time-profile}
\paragraph{StarVLA evaluator.}
Figure~\ref{fig:v2-native-timeline} profiles the StarVLA policy server and LIBERO
evaluator (revision \texttt{eaa51c4}, bf16) with the settings of the official LIBERO-Long evaluation (no video
logging, one BLAS thread) on LIBERO-Long, the first five initial states of each task (50 episodes),
seed 7, on one RTX 4090 with a 20-core quota; shares are means of three timed runs. A profiling wrapper instruments
the evaluator's environment creation, reset, initial-state loading, step, and close, the client's
action request, and the server's request handling, recording monotonic start and end times; each
step also accumulates MuJoCo physics time and records each rendering call. \texttt{nvidia-smi} samples GPU utilization
at 1\,Hz, and kernel time and launches per inference come from an Nsight Systems trace of the server
over ten episodes.

\paragraph{\system{} configurations.}
Table~\ref{tab:v2-time-profile} gives the stage-timed \system{} runs behind
Section~\ref{sec:v2-scalability}. The workload is the same 50 episodes on the same host, using the
checkpoint and Policy of the cross-system comparison. Each configuration runs three times.
Stage timing records monotonic start and end times in every process and marks the same ranges with
NVTX; per control step it accumulates MuJoCo physics, robot control and forward kinematics, observable
updates, and rendering. Nsight Systems traces CUDA kernels, NVTX ranges, and OS calls; py-spy samples
Python stacks in every process; PyTorch Profiler counts operators and kernel launches per inference;
NVML sampling gives GPU utilization.

\begin{table}[tbp]
\centering
\small
\caption{Simulator-worker time by stage (\% of worker wall time) and Policy-replica measurements,
mean of three runs.}
\label{tab:v2-time-profile}
\begin{tabular}{@{}l r r r@{}}
\toprule
 & $K1/g1$ & $K4/g1$ & $K12/g2$ \\
\midrule
Evaluation time (s) & 374.1 & 130.2 & 82.7 \\
\addlinespace
Robot control and forward kinematics & 37.1 & 28.6 & 16.0 \\
Physics integration & 10.4 & 7.9 & 4.4 \\
Observable updates and step bookkeeping & 2.9 & 2.3 & 1.3 \\
Rendering & 2.9 & 3.1 & 3.3 \\
Episode reset & 11.9 & 9.7 & 5.7 \\
Task environment creation & 3.2 & 4.1 & 6.0 \\
Settling steps & 1.8 & 1.4 & 0.9 \\
Waiting for actions & 29.4 & 41.7 & 60.0 \\
Episode RPCs and idle & 0.5 & 1.3 & 2.4 \\
\addlinespace
Mean GPU utilization (\%) & 15.4 & 42.3 & 80.3 \\
Replica busy time (\% per replica) & 26 & 85 & 85 \\
Request wait for a replica (ms) & 0 & 48 & 221 \\
Inference time (ms) & 55.6 & 63.5 & 84.3 \\
\bottomrule
\end{tabular}
\end{table}

Table~\ref{tab:v2-model-stages} splits one inference at $K1/g1$. Wall times come from a run that
synchronizes the GPU at each stage boundary; kernel time is the union of CUDA kernel intervals inside
each stage, measured under Nsight Systems. The model runs one forward pass of Qwen2.5-VL-3B and an MLP
action head, with no autoregressive decoding.

\begin{table}[tbp]
\centering
\small
\caption{One StarVLA LIBERO inference by model stage at $K1/g1$ (medians over 273 inferences after
20 warmup inferences).}
\label{tab:v2-model-stages}
\begin{tabular}{@{}l r r r r@{}}
\toprule
Stage & Wall (ms) & Kernel time (ms) & GPU busy (\%) & Kernels \\
\midrule
Image processing and tokenization & 4.7 & 0.0 & 0 & 0 \\
Vision encoder & 23.9 & 9.7 & 41 & 1,486 \\
Language model & 24.6 & 14.7 & 60 & 1,641 \\
Action head & 0.6 & 0.4 & -- & 30 \\
\midrule
Whole inference & 57.7 & 26.1 & 45 & 3,220 \\
\bottomrule
\end{tabular}
\end{table}

Each request from a worker to a replica carries about 525\,KB;
the worker encodes it in 2.6--2.7\,ms and the replica decodes it in 2.1--2.4\,ms. py-spy places the
simulation hot spots in MuJoCo forward kinematics, the physics step, the operational-space
controller, and model compilation from XML at episode reset.

\paragraph{\rev{RoboTwin Easy.}}
\rev{Table~\ref{tab:v2-robotwin-profile} profiles the StarVLA RoboTwin evaluator (\texttt{start\_eval.sh}, one
GPU) and \system{} on one host with four RTX PRO 5000 GPUs, over the same 100 RoboTwin Easy episodes: every fifth task
of the 500-episode list, all ten seeds of each, so the official instruction draw is unchanged. The same wrapper times
each task's scene setup, the official \rev{expert check}, each control step with its physics and rendering, observations
between steps, and each Policy round trip; the \system{} replica times each request under its request lock.
Each column is the mean of three timed runs; an untimed run of each differs in wall time by at most 1.7\%, and all
runs match the official evaluation on every success outcome and terminal step. The four official runs, three timed
and one untimed, ran in parallel, one per GPU.}

{\revcolor
\begin{table}[tbp]
\revcolor
\centering
\small
\caption{StarVLA-evaluator and simulator-worker time by stage on RoboTwin Easy (\% of evaluator or worker time), 100 episodes (all ten of every fifth task),
mean of three runs. The StarVLA evaluator runs on one GPU; \system{} on four RTX PRO 5000 GPUs.}
\label{tab:v2-robotwin-profile}
\begin{tabular}{@{}l r r r r@{}}
\toprule
 & Official & $K4/g4$ & $K32/g4$ & $K48/g8$ \\
\midrule
Wall time (s) & 4,971 & 565 & 256 & 297 \\
\addlinespace
Observations (rendering) & 43.5 & 14.3 & 45.0 & 60.9 \\
Controller and other step Python & 23.6 & 46.1 & 21.1 & 15.2 \\
Physics & 12.9 & 28.3 & 13.2 & 9.5 \\
Expert replay & 6.9 & -- & -- & -- \\
Scene setup & 5.2 & 1.6 & 1.7 & 2.4 \\
Waiting for actions & 4.2 & 9.4 & 18.7 & 12.0 \\
Other & 3.4 & 0.3 & 0.2 & 0.0 \\
\addlinespace
First observation of a new scene (s, median) & -- & 0.6 & 1.4 & 13.5 \\
CPU share of observation time (\%) & -- & 52 & 32 & 28 \\
Policy round trip / server time (ms, median) & -- & 82 / 73 & 342 / 127 & 281 / 133 \\
Replica busy time (\% per replica) & 3.9 & 8 & 35 & 19 \\
GPU time with a kernel running (\%) & 37 & 8 & 23 & 23 \\
Rendered observations never read (\%) & 93.7 & 0 & 0 & 0 \\
\bottomrule
\end{tabular}
\par\smallskip
\raggedright\footnotesize Restarting the policy server for each task adds a further 5.3\% to the StarVLA evaluator's wall time. \system{} renders only at chunk boundaries, so every rendered observation
reaches the Policy and \rev{the expert check} is replaced by the seed cache.\par
\end{table}
}

{\revcolor
In these 100-episode runs each \system{} worker evaluates about two episodes, so the first observations of new
processes and scenes weigh heavily. At $K4/g4$ the workers spend 74\% of their time in physics and the controller
on CPU cores. At $K48/g8$, observations take 61\% of worker time, of which only 28\% is CPU time, and the first
observation of each new scene takes 13.5\,s instead of 0.6\,s, while the replicas are busy only 19\% of the time. A
benchmark that only builds scenes and takes one observation, with no Policy, shows where that time goes: with 48
such processes, each Vulkan fence creation and destruction\rev{,} each Vulkan--CUDA buffer export\rev{, and each CUDA device-property query} of the ray-tracing
renderer and its denoiser takes 15--105\,ms instead of 0.3--2\,ms alone, while the host's CPU cores and GPUs stay
mostly idle. The calls queue in the GPU driver.

\paragraph{RoboTwin Easy stage shares over 500 episodes.}
Table~\ref{tab:v2-robotwin-stages} gives the stage shares of three 500-episode runs of the formal list at $K48/g8$
on four RTX PRO 5000 GPUs, from a launcher that times scene setup, every observation, every control step, and every
Policy request in each simulator worker, starting from each process's second episode; all three runs match the
official evaluation on every episode. Worker time goes 50\% to control steps (physics and the controller, on CPU
cores), 24\% to waiting for the Policy, 14\% to the first observation of each new scene, and 9\% to later
observations, so CPU time and GPU time both bound throughput. The waits come from sharing each GPU between two
replicas and twelve rendering processes: a Policy request takes 302\,ms \rev{on average} against \rev{a median of} 82\,ms at $K4/g4$, and an
observation of three cameras 121\,ms \rev{on average} against \rev{a median of} about 30\,ms for one process alone on a GPU (about 42\,ms for four, as in the official evaluator). Going
from $K36/g8$ to $K48/g8$, a third more workers, raises steady-state throughput by 13\%; adding 8\,ms to every
control step, a fifth more, lowers throughput by only 4.6\%, because requests and observations then wait less.
Rendering the three cameras the Policy reads, $K60/g8$ reaches 575 steady-state steps/s on the 5,000 episodes against 558 at $K48/g8$
(Table~\ref{tab:v2-4gpu-5000}). CUDA MPS,
which merges the CUDA contexts of the processes but not the Vulkan contexts that submit the ray tracing, adds
1.5--2\%. A per-GPU lock that lets one simulator worker render at a time leaves throughput unchanged (601 steps/s
against 595).

\begin{table}[tbp]
\revcolor
\centering
\small
\caption{\system{} simulator-worker time by stage on four RTX PRO 5000 GPUs, 500 RoboTwin Easy episodes, one run per
column, from the second episode of each simulator worker on, rendering the three cameras the Policy reads.}
\label{tab:v2-robotwin-stages}
\begin{tabular}{@{}l r r r@{}}
\toprule
 & $K48/g8$ & $K36/g8$ & \shortstack[r]{$K48/g8$, 8\,ms added\\to each control step} \\
\midrule
Steady-state throughput (steps/s) & 591 & 522 & 564 \\
\addlinespace
Control steps (\% of worker time) & 50.3 & 59.1 & 49.4 \\
Added wait (\%) & -- & -- & 9.4 \\
Policy requests (\%) & 23.5 & 18.9 & 19.1 \\
First observation of a new scene (\%) & 13.7 & 10.8 & 11.4 \\
Later observations (\%) & 9.2 & 8.6 & 8.0 \\
Scene setup and other (\%) & 3.2 & 2.6 & 2.8 \\
\addlinespace
Control step (ms) & 41.0 & 40.7 & 42.1 \\
Policy request (ms) & 302 & 206 & 257 \\
Later observation (ms) & 121 & 95 & 110 \\
First observation of a new scene (s, median) & 1.5 & 1.1 & 1.3 \\
\bottomrule
\end{tabular}
\end{table}
}

\section{Extended Results}
\label{app:v2-extended}
\rev{Each subsection below extends the subsection of Section~\ref{sec:v2-evaluation} with the same title.}
\subsection{Outcome Agreement}
\label{app:v2-outcome}
\paragraph{Official reference evaluations.}
The three official LIBERO-Long launches each score 451/500 (90.2\%) with identical success outcomes; per-task
success counts are 48, 50, 50, 46, 46, 49, 41, 42, 30, and 49. The fixed $K1/g1$ \system{} run of the
worker and replica sweep matches all 500 success outcomes of each launch, and its terminal steps
match on 500, 499, and 499 episodes. In the September batch (Appendix~\ref{app:v2-revisions}), the
official evaluation scores 439/500 (87.8\%) and a fixed $K1/g1$ \system{} run matches all 500 success
outcomes; per-task success counts are 50, 49, 50, 48, 42, 50, 37, 36, 30, and 47.

\paragraph{Per-episode outcome differences.}
On RoboTwin Easy, the stride-16 run on four RTX 4090 GPUs matches each of the three $K1/g1$ runs on one RTX 4090 on
every success outcome and terminal step, and the stride-1 runs on four RTX 4090 GPUs agree with one another
across $K4$ to $K32$ and 48 and 80 cores. On LIBERO-Long, Table~\ref{tab:v2-h1-scaling} gives the
success-outcome and terminal-step agreement of every worker and replica configuration with $K1/g1$.

\subsection{Performance}
\label{app:v2-performance}
\paragraph{\rev{Hosts and CPU quotas.}}
\label{app:v2-hosts}
\rev{Every LIBERO-Long cross-system launch runs on one RTX 4090 with 12 CPU cores, and every system runs three launches.
The worker and replica sweep and the three diagonal searches run on one RTX 4090 with a 20-core quota. RoboTwin
experiments run on one RTX 4090 for the $K1/g1$ runs and on four RTX 4090 GPUs with an 80-core quota. The RoboTwin
cross-system comparison runs one launch per system, one at a time, on one RTX 4090 with a 20-core quota. The RoboTwin
comparison with the official evaluation runs on one host with four RTX PRO 5000 GPUs and 208 CPU cores, one launch at
a time, alternating the official evaluation and \system{}.}
\paragraph{Search time.}
\label{app:v2-search-time}
The speedups in the main text compare evaluation time and leave out autotuning. On LIBERO-Long, the
search takes a median 506\,s with CUDA MPS off and 504\,s with it on, which adds 75\% to the evaluation time of
RoboScope Kit (Search); the grid it replaces takes eighteen launches and 5.4\,h of evaluation, against seven probes
and 8.4\,min for the search. On one RTX 4090, the RoboTwin Easy search takes 593\,s. On a 100-episode RoboTwin Easy
subset, the first two episodes of each task, the search takes about 60\% of each launch (Table~\ref{tab:v2-search-launches}). Tables~\ref{tab:v2-w1-search-runs}
and~\ref{tab:v2-search-launches} list the search time of every launch.
\paragraph{Per-run timings and repeats.}
The LIBERO-Long experiments use two RTX 4090 machines, host~A and host~B. StarVLA and RLinf ran on host~A with a 12-core CPU quota, launched in interleaved order on the
same GPU (Table~\ref{tab:v2-w1-runs}); the speedup of RLinf over StarVLA is the geometric mean of the
three paired ratios, with a 95\% $t$ interval over the log ratios. RoboScope Kit (Search) ran on another
RTX 4090 of host~A, and vla-eval on host~B, both with all processes pinned to 12 cores;
their speedups are ratios of geometric-mean evaluation times with a 95\% Welch $t$ interval on the log
times. The worker and replica sweep and the autotuning launches ran on host~B with a 20-core
quota. Success is the mean over launches; time and throughput are medians.

Table~\ref{tab:v2-c17-comparison} lists the values behind Figure~\ref{fig:v2-w1-results},
Table~\ref{tab:v2-w1-search-runs} the \system{} launches at $K12/g3$ with their CPU and GPU use,
Table~\ref{tab:v2-w1-runs} the StarVLA and RLinf launches, and Table~\ref{tab:v2-w1-best-runs} the vla-eval
launches and \system{} at a fixed $K8/g2$.

\newcommand{\withci}[3]{$#1$\;{\scriptsize\color{black!60}$[#2,\,#3]$}}

\begin{table}[tbp]
\begingroup
\footnotesize
\setlength{\tabcolsep}{2pt}
\renewcommand{\arraystretch}{1.16}
\caption{Speed and outcome agreement in the LIBERO-Long cross-system comparison.}
\label{tab:v2-c17-comparison}
\begin{tabularx}{\linewidth}{@{}l >{\centering\arraybackslash}X c *{4}{>{\centering\arraybackslash}X}@{}}
\toprule
System & \shortstack{Success\\(\%)} & \shortstack{$\Delta$ success vs.\\StarVLA, pp [95\% CI]} & \shortstack{Agreement with\\StarVLA (\%)} & \shortstack{Eval.\\time (s)} & \shortstack{Throughput\\(steps/s)} & \shortstack{Speedup\\vs. StarVLA}\\
\midrule
StarVLA official & 90.2 & \multicolumn{2}{c}{reference} & 2,990 & 45.8 & 1.0 \\
RLinf & 90.3 & \withci{+0.1}{-2.6}{+3.3} & 88.1 & 1,882 & 74.2 & 1.6 \\
vla-eval & 89.8 & \withci{-0.4}{-1.6}{+0.7} & 91.7 & 1,016 & 133.5 & 2.9 \\
\shortstack[l]{\textbf{RoboScope Kit}\\\textbf{(Search)}} & \textbf{89.9} & \withci{-0.3}{-0.8}{0.0} & 97.9 & \textbf{671} & \textbf{204.6} & \textbf{4.4} \\
\shortstack[l]{\textbf{RoboScope Kit}\\\textbf{(Search+MPS)}} & \textbf{89.9} & \withci{-0.3}{-1.3}{+0.8} & 96.5 & \textbf{485} & \textbf{282.5} & \textbf{6.2} \\
\bottomrule
\end{tabularx}
\par\smallskip
\raggedright StarVLA on LIBERO-Long, 500 episodes (10 tasks $\times$ 50, seed 7), one RTX 4090; StarVLA and RLinf each ran three launches in interleaved order on the same GPU with a 12-core CPU quota, vla-eval ran three launches at its tuned configuration on host~B with 12 pinned cores, and RoboScope Kit (Search) ran three launches on another GPU of host~A with 12 pinned cores, using the fixed-configuration runner of an earlier \system{} build with the same Policy computation and checkpoint. The official evaluation uses StarVLA's policy server and LIBERO evaluator. RoboScope Kit (Search) evaluates $K12/g3$, the configuration that autotuning in \system{} selected in all three of its launches on host~B (median search 506\,s, Table~\ref{tab:v2-w1-search-runs}). RoboScope Kit (Search+MPS) ran three launches of that search with CUDA MPS on another GPU of host~A with 12 pinned cores; each searched (median 504\,s) and evaluated the selected $K16/g4$. Success is the mean over launches; time and throughput are medians. $\Delta$ success is the row\textquotesingle{}s success rate minus the official StarVLA evaluation\textquotesingle{}s, with a 95\% paired task-cluster bootstrap interval (10 tasks, 20,000 draws); agreement is the share of the 500 episodes with the same success outcome as the official evaluation, launch $i$ paired with official launch $i$. Speedup is StarVLA evaluation time / row evaluation time: for RLinf the geometric mean over the three launches, launch $i$ paired with StarVLA launch $i$; for vla-eval and both RoboScope Kit rows the ratio of geometric-mean evaluation times. \rev{Each speedup thus uses all three launches of both systems, while the time column shows the median launch.} Throughput is policy-controlled environment steps divided by evaluation time, excluding settling.\par
\endgroup
\end{table}

\begin{table}[tbp]
\centering
\footnotesize
\setlength{\tabcolsep}{3.5pt}
\renewcommand{\arraystretch}{1.12}
\caption{RoboScope Kit (Search) and (Search+MPS) launches, with CPU and GPU use during the formal evaluation.}
\label{tab:v2-w1-search-runs}
\begin{tabular}{@{}l c r r r r r r r r@{}}
\toprule
Launch & MPS & Successes & Search (s) & Eval.\ time (s) & \shortstack{Launch incl.\\search (s)} & Steps/s & \shortstack{CPU\\cores} & \shortstack{GPU util.\\mean (\%)} & \shortstack{Peak GPU\\mem.\ (GiB)} \\
\midrule
Search 1, host~B & Off & 448 & 506.0 & 679.0 & 1,192.6 & 202.5 & 7.8 & 97.0 & 33.6 \\
Search 2, host~B & Off & 447 & 504.1 & 675.2 & 1,187.0 & 204.6 & 7.8 & 97.2 & 33.6 \\
Search 3, host~B & Off & 448 & 506.9 & 679.7 & 1,193.9 & 202.5 & 7.8 & 96.8 & 33.5 \\
\midrule
Fixed 1, host~A & Off & 449 & -- & 671.2 & 737.6 & 204.9 & 7.7 & 96.8 & 33.8 \\
Fixed 2, host~A & Off & 448 & -- & 673.5 & 739.1 & 204.6 & 7.8 & 96.6 & 33.7 \\
Fixed 3, host~A & Off & 451 & -- & 669.3 & 731.0 & 204.6 & 7.8 & 96.6 & 33.7 \\
\midrule
Search+MPS 1, host~A & On & 452 & 503.9 & 484.6 & 997.1 & 281.8 & 10.6 & 95.0 & 44.7 \\
Search+MPS 2, host~A & On & 447 & 504.8 & 486.9 & 1,000.5 & 282.5 & 10.6 & 95.6 & 44.5 \\
Search+MPS 3, host~A & On & 449 & 502.2 & 483.4 & 994.2 & 283.3 & 10.5 & 95.8 & 44.4 \\
\bottomrule
\end{tabular}
\par\smallskip
\raggedright The three searches are the LIBERO-Long launches of Table~\ref{tab:v2-search-launches}: one RTX 4090 and a 20-core quota; each searches and then evaluates the selected $K12/g3$. The fixed launches evaluate $K12/g3$ without search on host~A with all processes pinned to 12 cores, using the fixed-configuration runner of an earlier \system{} build; they are the RoboScope Kit (Search) row of Table~\ref{tab:v2-c17-comparison}. The Search+MPS launches run the diagonal search with CUDA MPS on host~A with 12 pinned cores, select $K16/g4$, and are the RoboScope Kit (Search+MPS) row. CPU cores are the mean cores the container consumed during the formal evaluation, from cgroup CPU time; CPU throttling stays below 0.7\% of the quota. GPU utilization is the mean of 1\,Hz NVIDIA samples during the formal evaluation, and peaks at 100\% in every launch. Every launch passes the LIBERO-Long audit; the MPS launches observe four MPS clients. Appendix~\ref{app:v2-time-profile} profiles the CPU and GPU use of the StarVLA evaluator.\par
\end{table}

\begin{table}[tbp]
\centering
\footnotesize
\setlength{\tabcolsep}{5pt}
\renewcommand{\arraystretch}{1.12}
\caption{StarVLA and RLinf launches in execution order (same host and GPU).}
\label{tab:v2-w1-runs}
\begin{tabular}{@{}r l c r r r r@{}}
\toprule
Order & System & Launch & Successes & Eval.\ time (s) & End-to-end (s) & Steps/s \\
\midrule
1 & StarVLA & 1 & 451 & 2,989.9 & 3,116.5 & 45.8 \\
2 & RLinf & 1 & 453 & 1,881.7 & 1,934.8 & 74.2 \\
3 & RLinf & 2 & 452 & 1,907.5 & 1,955.3 & 72.0 \\
4 & StarVLA & 2 & 451 & 2,968.0 & 2,983.5 & 46.1 \\
5 & RLinf & 3 & 450 & 1,806.7 & 1,857.8 & 77.3 \\
6 & StarVLA & 3 & 451 & 2,997.9 & 3,012.5 & 45.6 \\
\bottomrule
\end{tabular}
\par\smallskip
\raggedright End-to-end is the monotonic wall time of the whole process. RLinf was tuned once offline (1,704.3\,s); including that tuning, its median end-to-end time is 3,639\,s.\par\end{table}

\begin{table}[tbp]
\centering
\footnotesize
\setlength{\tabcolsep}{5pt}
\renewcommand{\arraystretch}{1.12}
\caption{Launches of \system{} at a fixed $K8/g2$, the fastest configuration of the sweep with two replicas, and of vla-eval at its tuned configuration, in execution order.}
\label{tab:v2-w1-best-runs}
\begin{tabular}{@{}r l r r r r@{}}
\toprule
Order & System & Successes & Eval.\ time (s) & End-to-end (s) & Steps/s \\
\midrule
1 & \system{} $K8/g2$ & 450 & 762.2 & 805.5 & 179.1 \\
2 & \system{} $K8/g2$ & 451 & 744.4 & 785.5 & 183.4 \\
3 & \system{} $K8/g2$ & 445 & 756.4 & 804.4 & 181.3 \\
4 & \system{} $K8/g2$ & 448 & 754.6 & 841.5 & 181.8 \\
5 & \system{} $K8/g2$ & 449 & 745.8 & 789.2 & 183.2 \\
6 & vla-eval & 446 & 1,007.0 & 1,023.6 & 135.1 \\
7 & vla-eval & 451 & 1,016.2 & 1,041.8 & 133.5 \\
8 & vla-eval & 450 & 1,036.0 & 1,055.1 & 132.8 \\
\bottomrule
\end{tabular}
\par\smallskip
\raggedright vla-eval runs the configuration selected in Table~\ref{tab:v2-vla-tuning}; its offline tuning took 9,780.1\,s, and including it, median end-to-end time is 10,822\,s. Every launch passes the LIBERO-Long audit: 500 unique episodes with the frozen initial states, seed 7, and native outcomes equal to the observer's.\par
\end{table}

\subsection{Ablation Study}
\label{app:v2-ablation}
\paragraph{Fixed-configuration runs.}
Seven initial runs use fixed configurations: two LIBERO-Long $K1/g1$ runs on a
12-core host, three RoboTwin Easy $K1/g1$ runs on one RTX 4090, and two RoboTwin Easy $K4/g4$ runs on four
RTX 4090 GPUs with a 48-core quota. Both LIBERO-Long runs score 87.8\% with identical success outcomes. RoboTwin Easy scores
46.2\% in all three runs on one GPU and 47.0\% in both runs on four, with success outcomes and
terminal steps identical within each configuration (Table~\ref{tab:v2-fixed-runs}).

\begin{table}[tbp]
\centering
\small
\caption{Initial fixed-configuration runs. Each run has 500 episodes. Timing and throughput are means
$\pm$ sample standard deviations; $n$ counts runs.}
\label{tab:v2-fixed-runs}
\begin{tabular}{@{}l c r r r r@{}}
\toprule
Workload & $K/g$ & $n$ & Success/run & Eval.\ time (s) & Steps/s \\
\midrule
LIBERO-Long & 1/1 & 2 & 439/500 & $3{,}069 \pm 54$ & $45.49 \pm 0.81$ \\
RoboTwin Easy & 1/1 & 3 & 231/500 & $21{,}582 \pm 389$ & $9.79 \pm 0.17$ \\
RoboTwin Easy & 4/4 & 2 & 235/500 & $5{,}646 \pm 193$ & $36.86 \pm 1.26$ \\
\bottomrule
\end{tabular}
\end{table}

\paragraph{Autotuning launches.}
Autotuning runs on one RTX 4090 with a 20-core quota, without CPU pinning or CUDA MPS,
three times per workload. Each launch is one process that searches and then evaluates at the selected
$K/g$. LIBERO-Long uses the frozen 500-episode manifest; the RoboTwin Easy subset takes the first two
episodes of each of the 50 tasks, with observation stride 16 and a 60-minute search budget. An audit
checks that the formal $K/g$ equals the selection, that no worker fails or runs out of memory, and that
the episode records cover exactly the manifest; on LIBERO-Long, an observer additionally records every
formal episode and checks its initial state, outcome, and step count against the manifest and the
native records. \rev{All six launches pass every check that applies to them.} In every LIBERO-Long search without MPS, the worker phase reaches the knee at $K4/g1$:
doubling to $K8/g1$ gains only 0.4--9.6\%, under 25\%. The replica phase then stops at $K12/g3$:
$K16/g4$ gains only 2.1--2.4\%, under 3\%. Three further LIBERO-Long launches run the search with CUDA MPS
on host~A with all processes pinned to 12 cores; all pass the same audit. Their worker
phase also reaches the knee at $K4/g1$ ($K8/g1$ gains 7.6--8.6\%), their replica phase accepts $K16/g4$
(8.2--9.4\% over $K12/g3$), and $K20/g5$ is skipped because its predicted memory, 54.7--55.2\,GiB,
exceeds the 46\,GiB usable on the GPU. Table~\ref{tab:v2-mps-grid} lists the fixed-configuration grid
under MPS; its $K16/g4$ launch evaluates in 482.5\,s with 450 successes, within 0.5\%
of the searched launches. In every RoboTwin search, the $K16/g1$ probe runs out of GPU memory, so the search keeps
$K8/g1$; the three RoboTwin launches match on every success outcome and terminal step.

\begin{table}[tbp]
\centering
\footnotesize
\setlength{\tabcolsep}{4pt}
\caption{Autotuning launches on one RTX 4090, with a 20-core quota or, with CUDA MPS, 12 pinned cores
on host~A. Evaluation time starts after the search; the launch time is the wall time of the whole launch, search included.}
\label{tab:v2-search-launches}
\begin{tabular}{@{}l c r r r r r@{}}
\toprule
Workload & $K/g$ & Search (s) & Eval.\ time (s) & \shortstack{Launch incl.\\search (s)} & Successes & \shortstack{Peak GPU\\mem.\ (GiB)} \\
\midrule
LIBERO-Long & 12/3 & 506 & 679 & 1,193 & 448/500 & 33.6 \\
 & 12/3 & 504 & 675 & 1,187 & 447/500 & 33.6 \\
 & 12/3 & 507 & 680 & 1,194 & 448/500 & 33.5 \\
\midrule
LIBERO-Long, & 16/4 & 504 & 485 & 997 & 452/500 & 44.7 \\
CUDA MPS, & 16/4 & 505 & 487 & 1,000 & 447/500 & 44.5 \\
12 pinned cores & 16/4 & 502 & 483 & 994 & 449/500 & 44.4 \\
\midrule
RoboTwin Easy & 8/1 & 538 & 343 & 884 & 52/100 & 41.0 \\
(50 tasks $\times$ 2) & 8/1 & 561 & 351 & 915 & 52/100 & 40.7 \\
 & 8/1 & 616 & 351 & 986 & 52/100 & 41.5 \\
\bottomrule
\end{tabular}
\end{table}

\begin{table}[tbp]
\centering
\footnotesize
\setlength{\tabcolsep}{4pt}
\caption{LIBERO-Long at fixed $K/g$ under CUDA MPS with 12 pinned cores, one
launch per configuration (the grid of Figure~\ref{fig:v2-scaling}b).}
\label{tab:v2-mps-grid}
\begin{tabular}{@{}c r r r r r r r@{}}
\toprule
$K/g$ & Successes & Eval.\ time (s) & End-to-end (s) & Steps/s & \shortstack{CPU\\cores} & \shortstack{GPU util.\\mean (\%)} & \shortstack{Peak GPU\\mem.\ (GiB)} \\
\midrule
1/1 & 451 & 3,655.4 & 3,676.2 & 37.4 & 1.2 & 15.0 & 9.6 \\
2/1 & 453 & 1,908.3 & 1,932.5 & 71.6 & 2.2 & 29.5 & 10.5 \\
4/1 & 450 & 1,177.9 & 1,204.5 & 115.9 & 3.4 & 48.5 & 11.5 \\
8/1 & 451 & 1,166.5 & 1,190.5 & 117.2 & 3.5 & 49.5 & 13.5 \\
12/1 & 448 & 1,171.4 & 1,192.2 & 117.6 & 3.6 & 49.8 & 16.1 \\
16/1 & 451 & 1,186.3 & 1,208.9 & 115.5 & 3.5 & 49.0 & 18.0 \\
20/1 & 444 & 1,170.3 & 1,193.6 & 117.2 & 3.6 & 49.6 & 19.6 \\
\midrule
4/2 & 451 & 992.0 & 1,033.2 & 138.2 & 4.1 & 53.8 & 20.4 \\
8/2 & 450 & 637.3 & 693.1 & 214.2 & 6.6 & 79.3 & 22.3 \\
12/2 & 449 & 623.4 & 672.1 & 220.5 & 6.9 & 81.3 & 25.2 \\
16/2 & 451 & 621.2 & 666.5 & 219.8 & 6.9 & 81.2 & 27.0 \\
\midrule
8/3 & 449 & 581.2 & 638.8 & 234.8 & 7.5 & 84.6 & 31.1 \\
12/3 & 448 & 518.9 & 588.2 & 265.6 & 9.1 & 92.7 & 33.6 \\
16/3 & 447 & 513.7 & 582.7 & 266.7 & 9.1 & 92.9 & 35.1 \\
20/3 & 450 & 511.2 & 575.7 & 266.8 & 9.1 & 92.8 & 37.4 \\
\midrule
12/4 & 446 & 489.2 & 574.4 & 281.9 & 10.2 & 95.4 & 42.8 \\
\textbf{16/4} & 450 & \textbf{482.5} & 573.7 & 284.3 & 10.6 & 96.0 & 44.4 \\
20/4 & 442 & 483.2 & 575.1 & 284.9 & 10.5 & 95.7 & 46.3 \\
\bottomrule
\end{tabular}
\par\smallskip
\raggedright Each launch evaluates the 500 frozen episodes with the fixed-configuration runner of the earlier \system{} build used for the RoboScope Kit (Search) row, and passes the LIBERO-Long audit. The $g=1$ and $K4/g2$ launches ran later on a second machine configured like host~A, with the same runner and pinning. The diagonal search under MPS selects $K16/g4$ (Table~\ref{tab:v2-search-launches}). CPU cores and GPU utilization are means during the formal evaluation.\par
\end{table}

\paragraph{Controlled Simulation Runtime ablations.}
The ladder runs one RoboTwin task with three seeds on one RTX 4090; each variant adds one Simulation Runtime
change to the previous one and runs once, with a Policy that holds the current joint state in place of
a model.

\begin{table}[tbp]
\centering
\small
\caption{Cumulative Simulation Runtime ladder on one RTX 4090. The Policy holds the current joint state in place of a
model. Each row adds one change to the row above and runs once, on one RoboTwin task with three seeds. Steps/s divides
the 2,100 policy steps by wall time.}
\label{tab:v2-e18-ladder}
\begin{tabularx}{\linewidth}{@{}>{\raggedright\arraybackslash}p{28mm} X r r r r@{}}
\toprule
Variant & Added Simulation Runtime behavior & Observations & Setup (s) & Wall (s) & Steps/s \\
\midrule
00 official-like & Official-like path & 2,100 & 46.9 & 237 & 8.88 \\
01 shared context & Reuse Engine and Renderer & 2,100 & 45.2 & 226 & 9.31 \\
02 planner warmup & Skip cuRobo initialization & 2,100 & 1.4 & 180 & 11.63 \\
03 chunk boundary & Observe at 16-step chunk boundaries & 132 & 1.4 & 71.9 & 29.23 \\
04 passive force & Hoist duplicate passive force & 132 & 1.7 & 64.8 & 32.38 \\
\bottomrule
\end{tabularx}
\end{table}

\paragraph{\rev{Four RTX 4090 GPUs.}}
\begin{table}[tbp]
\centering
\small
\caption{Evaluation time of the 500 RoboTwin Easy episodes with the real StarVLA Policy, along the measured
configurations \rev{on RTX 4090 GPUs, starting from \system{}'s single-GPU $K1/g1$ baseline; every row reuses the Engine and Renderer and hoists the duplicate passive force}. Each factor divides the previous row's time by the current row's.}
\label{tab:v2-4090-chain}
\begin{tabular}{@{}l l r r r@{}}
\toprule
Change & Configuration & Runs & Eval.\ time (s) & Factor \\
\midrule
Baseline & $K1/g1$, one GPU, stride 1 & 3 & 21,582 & -- \\
Four GPUs & $K4/g4$, four GPUs, 80 cores & 3 & 5,934 & 3.64$\times$ \\
Workers & $K16/g4$, same hardware & 1 & 3,951 & 1.50$\times$ \\
Stride and run setup & $K16/g4$, stride 16 & 1 & 862 & 4.58$\times$ \\
\midrule
Total & & & & 25.0$\times$ \\
\bottomrule
\end{tabular}
\end{table}

\begin{table}[tbp]
\centering
\small
\caption{Timing and outcome of the RoboTwin Easy stride-16 run ($K16/g4$, four RTX 4090, 80 cores).}
\label{tab:v2-stride16}
\begin{tabular}{@{}l r@{}}
\toprule
Measurement & Value \\
\midrule
Evaluation time & 862\,s \\
Metrics-recorder time & 855\,s \\
Policy loading & 65.8\,s \\
Runtime preparation & 0.2\,s \\
End-to-end time & 928\,s \\
Successes/episodes & 231/500 \\
Total action steps & 211,220 \\
OOM events/worker failures & 0/0 \\
\bottomrule
\end{tabular}
\end{table}
The run uses $K16/g4$ and observation stride 16. Its setup uses work items of two episodes instead of ten, and
loads RoboTwin's static clutter-object metadata from a cache instead of rebuilding it in each simulator worker.
Against each of the
three $K1/g1$ runs on one RTX 4090, all 500 episode keys, success outcomes, and terminal steps match. Its evaluation
time is 6.9$\times$ lower than the fixed $K4/g4$ mean of Table~\ref{tab:v2-4090-chain}.

{\revcolor
\paragraph{Four RTX PRO 5000 GPUs.}
The official evaluation and \system{} each ran three launches on one host with four RTX PRO 5000 GPUs and 208 CPU
cores, in alternating order, with the official StarVLA RoboTwin checkpoint, RoboTwin Easy with the 500 seeds of
the comparison on one RTX 4090, and one asset directory for both systems. Table~\ref{tab:v2-4gpu-runs} lists each launch.
Table~\ref{tab:v2-speedup-chain} steps from the official evaluation to the fastest measured \system{}
configuration; Table~\ref{tab:v2-pro5000-ladder} repeats the Simulation Runtime ladder with the real Policy; and
Table~\ref{tab:v2-pro5000-scaling} scales from one GPU to four. \rev{In three launches with an earlier version of the search, it} takes a mean 1,574\,s, 69\% of the
2,284\,s from launch to exit, and selects $K32/g4$ each time. The three launches of the official evaluation, and
the three of \system{} with its search, are each within 1.3\% of their mean wall time. The first fixed $K48/g8$ launch with the longest tasks first took 675\,s
against 594 and 602\,s\rev{, most of the difference in replica loading}. At $K64/g8$ the renderer runs out
of GPU memory: a 731\,s run had five such failures and seven episodes that differ from the official evaluation.

The fixed $K48/g8$ uses one-episode work items and replicas on different GPUs started in parallel; the task-order
ablation below dispatches the longest tasks first. \rev{With the current search} and a run cache, one launch of
the default flow searched for 1,558\,s, selected $K60/g8$, evaluated the 500 episodes in 706\,s, stored the
configuration with the mean episode time of each task, and took 2,264\,s from launch to exit; a second
launch read the stored entry, ran no probes, dispatched the longest tasks first, and took 621\,s. Both match the
official evaluation on all 500 episodes; $K60/g8$ peaks at 67.8--70.2\,GiB of the 71.7\,GiB of each GPU. On the
standard 5,000 episodes in the default order, $K60/g8$ takes 3,985\,s of evaluation, against 4,023\,s at $K48/g8$,
and matches the official evaluation on every episode (Table~\ref{tab:v2-4gpu-5000}).

\begin{table}[tbp]
\revcolor
\centering
\small
\caption{Launches on four RTX PRO 5000 GPUs. Wall time is from launch to exit. \system{} evaluation time counts from
the end of replica loading, after the search.}
\label{tab:v2-4gpu-runs}
\begin{tabular}{@{}c r r r r r@{}}
\toprule
Launch & Official (s) & \system{} (s) & Search (s) & \system{} eval.\ time (s) & $K48/g8$ longest first (s) \\
\midrule
1 & 7,156 & 2,260 & 1,543 & 713 & 675 \\
2 & 7,131 & 2,313 & 1,615 & 695 & 594 \\
3 & 7,046 & 2,279 & 1,564 & 711 & 602 \\
\midrule
Mean & 7,111 & 2,284 & 1,574 & 706 & 624 \\
\bottomrule
\end{tabular}
\end{table}

\begin{table}[tbp]
\revcolor
\centering
\small
\caption{Evaluation time of the 500 RoboTwin Easy episodes on four RTX PRO 5000 GPUs, along the measured configurations,
measured as in Table~\ref{tab:v2-4gpu-headline}. Each factor divides the previous row's time by the current row's. Work items are episodes per queue item.}
\label{tab:v2-speedup-chain}
\begin{tabularx}{\linewidth}{@{}l X r r r@{}}
\toprule
Change & Configuration & Runs & Eval.\ time (s) & Factor \\
\midrule
StarVLA official & One evaluator per GPU & 3 & 7,076 & -- \\
\system{} architecture & $K4/g4$, 10-episode items, runtime optimizations off & 1 & 5,904 & 1.20$\times$ \\
Runtime optimizations & \rev{Engine and Renderer reuse, skip cuRobo initialization, hoist duplicate passive force}, stride 16 & 1 & 2,827 & 2.09$\times$ \\
Workers & $K16/g4$, 5-episode items & 1 & 970 & 2.91$\times$ \\
Smaller items & $K16/g4$, 2-episode items & 1 & 890 & 1.09$\times$ \\
Workers & $K32/g4$, 2-episode items & 1 & 694 & 1.28$\times$ \\
Replica start & $K32/g4$, 1-episode items, parallel replica start across GPUs & 1 & 683 & 1.02$\times$ \\
Workers and replicas & $K48/g8$ & 1 & 627 & 1.09$\times$ \\
Longest tasks first & $K48/g8$ & 3 & 573 & 1.09$\times$ \\
\midrule
Total & & & & 12.3$\times$ \\
\bottomrule
\end{tabularx}
\end{table}

\begin{table}[tbp]
\revcolor
\centering
\small
\caption{Cumulative Simulation Runtime ladder with the real StarVLA Policy on four RTX PRO 5000 GPUs ($K4/g4$),
100 RoboTwin Easy episodes, the first two of each of the 50 tasks. Each row adds one change to the row above and runs once. All rows
have 47 successes and 41,301 policy steps, and match the official evaluation episode by episode. Evaluation time
starts at the end of replica loading.}
\label{tab:v2-pro5000-ladder}
\begin{tabular}{@{}l r r r r@{}}
\toprule
Added Simulation Runtime behavior & Eval.\ time (s) & Steps/s & Factor & Cumulative \\
\midrule
None (render every step) & 1,349 & 30.6 & -- & 1.0$\times$ \\
Reuse Engine and Renderer & 1,274 & 32.4 & 1.06$\times$ & 1.1$\times$ \\
\rev{Skip cuRobo initialization} & 1,106 & 37.3 & 1.15$\times$ & 1.2$\times$ \\
\rev{Hoist duplicate passive force} & 1,067 & 38.7 & 1.04$\times$ & 1.3$\times$ \\
Observe at 16-step chunk boundaries & 552 & 74.8 & 1.93$\times$ & 2.4$\times$ \\
\bottomrule
\end{tabular}
\end{table}

\begin{table}[tbp]
\revcolor
\centering
\small
\caption{Throughput from one to four RTX PRO 5000 GPUs, 250 RoboTwin Easy episodes (5 per task), one-episode
work items. Steps/s counts the policy steps of the episodes that finish between the first and the third quarter
of the run's 250 completions, over the time between those two points. All six runs have 113/250 successes.}
\label{tab:v2-pro5000-scaling}
\begin{tabular}{@{}l r r@{}}
\toprule
Configuration & Steps/s & vs.\ one GPU \\
\midrule
One GPU, $K8/g1$ & 97 & 1.0$\times$ \\
One GPU, $K16/g1$ & 100 & 1.0$\times$ \\
Four GPUs, $K16/g4$ & 281 & 2.9$\times$ \\
Four GPUs, $K32/g4$ & 389 & 4.0$\times$ \\
Four GPUs, $K32/g8$ & 459 & 4.7$\times$ \\
Four GPUs, $K48/g8$ & 505 & 5.2$\times$ \\
\bottomrule
\end{tabular}
\end{table}

\paragraph{Standard 5,000-episode protocol.}
The official evaluation ran once, on an episode list built like that of the 500-episode comparison: the seeds of
\system{}'s seed generation for all 100 episodes of each task, and instructions drawn by the official rule over the
first 100 seeds of each task. Summed over episodes it takes 246,224\,s, 68 hours on one GPU, of which its expert
check takes 22,317\,s (9.1\%). Four RTX PRO 5000 GPUs bring the evaluation to 17.5 hours, and in the last hour only some GPUs
still run the longest tasks.
Table~\ref{tab:v2-4gpu-5000-tasks} lists each task's successes, which are equal in the official evaluation and all
four \system{} launches of Table~\ref{tab:v2-4gpu-5000}.

\paragraph{Task order.}
\system{}'s workers take one-episode work items from one queue in manifest order, which cycles through the tasks:
episode $i$ of every task, then episode $i{+}1$. Once a run has stored the mean episode time of every task, a later
run can instead dispatch the tasks longest first. This changes no episode's outcome or terminal step; it changes
only when each episode starts. Table~\ref{tab:v2-task-order} compares the two orders at $K48/g8$.

The order matters at the end of a run. While the queue has items, every worker is busy, so the evaluation phase
cannot be shorter than the summed episode time divided by the 48 workers. Once the queue empties, each worker
finishes the episode it holds and then idles, and the run ends when the last one finishes. RoboTwin Easy episode
times range widely: over the 5,000 episodes the median is 26.7\,s and the longest 207\,s, and the long episodes
come from a few tasks such as \code{blocks\_ranking\_size}, whose 100 episodes take the official evaluation
10,013\,s against 787\,s for \code{click\_bell} (Table~\ref{tab:v2-4gpu-5000-tasks}). In the default order,
episodes of the long tasks are still dispatched near the end: the last 48 items take 49\,s on average and up to
123\,s, and a worker idles 63\,s on average before the run ends. With the longest tasks first, the last items are
the short tasks: the last 48 take 18\,s on average and at most 53\,s, and a worker idles 7\,s. This is the
longest-processing-time-first rule for scheduling jobs on identical machines~\citep{graham1969bounds}.

Replaying the measured episode times of the 5,000-episode launches through a 48-worker queue isolates this end
effect, since it keeps each episode's time fixed and changes only the order (Table~\ref{tab:v2-task-order}, lower
half). Over the bound of 3,382\,s, the default order adds 64\,s, the longest tasks first 8\,s, and sorting
individual episodes longest first 4\,s, so ordering by task, which a stored per-task mean allows, gets nearly all
of the gain. Random task orders add 31\,s on average, and the shortest tasks first add 147\,s.

The measured difference is larger than the replay, for two reasons. In the default order a worker changes task
between 99\% of its consecutive episodes, against 47\% with the longest tasks first, and each change sets up a new
scene: the 5,000 episodes take 5,139 worker-seconds more in total, 107\,s across the 48 workers. Of the 238\,s
between the two 5,000-episode launches, 172\,s is in evaluation and 65\,s in replica loading (94\,s
against 29\,s), which does not depend on the order. On the 500 episodes, evaluation takes 627\,s in the
default order and 573\,s on average with the longest tasks first (566--579\,s over three launches); wall time is
655\,s against 624\,s.

\begin{table}[tbp]
\revcolor
\centering
\small
\caption{Task-order ablation at fixed $K48/g8$ on four RTX PRO 5000 GPUs. Upper half: measured launches; the 500-episode
longest-first row is the mean of three launches, the others ran once. Every launch matches the official evaluation
on every success outcome and terminal step. Lower half: the measured episode times of the two 5,000-episode launches
(the smaller of each episode's two times) replayed through a 48-worker queue; the time over the bound is the end of
the last episode minus the summed episode time divided by 48 (3,382\,s). The replay counts episode time only; the measured launches also include the cost of
changing task.}
\label{tab:v2-task-order}
\begin{tabular}{@{}l r r r r@{}}
\toprule
\multicolumn{5}{@{}l}{\emph{Measured launches}} \\
Episodes, task order & Wall (s) & Eval.\ time (s) & Idle at end (s) & Task changes \\
\midrule
500, default & 655 & 627 & -- & -- \\
500, longest tasks first & 624 & 573 & -- & -- \\
5,000, default & 4,119 & 4,023 & 63 & 99\% \\
5,000, longest tasks first & 3,881 & 3,851 & 7 & 47\% \\
\midrule
\multicolumn{5}{@{}l}{\emph{Replay of the 5,000-episode times}} \\
Order & \multicolumn{4}{r@{}}{Time over the bound (s)} \\
\midrule
Longest episodes first & \multicolumn{4}{r@{}}{4} \\
Longest tasks first, by task mean & \multicolumn{4}{r@{}}{8} \\
Default (episode $i$ of every task) & \multicolumn{4}{r@{}}{64} \\
Tasks in random order, mean of 2,000 & \multicolumn{4}{r@{}}{31 (5--95\%: 15--54)} \\
Episodes in random order, mean of 2,000 & \multicolumn{4}{r@{}}{74 (5--95\%: 46--112)} \\
Shortest tasks first & \multicolumn{4}{r@{}}{147} \\
\bottomrule
\end{tabular}
\end{table}

\paragraph{Front camera.}
RoboTwin renders four cameras for every observation and the Policy reads three. With the longest tasks first,
the $K48/g8$ launch of Table~\ref{tab:v2-4gpu-5000} that renders only those three reaches 569 steady-state steps/s,
against 511 with all four rendered; the head and wrist images and the joint state are bit-identical over 4 seeds and
about 31 observations each.

\paragraph{Scene memory.}
Over a run this long, the scenes of finished tasks stay on the GPU: the last minutes reach 71.1 of the 71.7\,GiB of
each GPU, and some launches have renderer OOM failures. \system{}'s RoboTwin launcher frees each finished task's
scene, and the per-GPU peak then stays at 59.7--63.7\,GiB at $K48/g8$ and 68.9--69.3\,GiB at $K60/g8$, with no OOM
failures.

\begin{table}[tbp]
\revcolor
\centering
\footnotesize
\setlength{\tabcolsep}{4pt}
\caption{Launches of the standard 5,000-episode protocol on four RTX PRO 5000 GPUs. The \system{} launches run
$K48/g8$, or $K60/g8$ where marked, with one-episode work items and parallel replica start, render the head and wrist cameras and, where
listed, the front camera, and free each finished task's scene; each matches the official evaluation on all 5,000 success outcomes and
terminal steps, with 2,313 successes. Wall time is from launch to exit; evaluation time and the ratio to the official
evaluation are measured as in Table~\ref{tab:v2-4gpu-headline}.}
\label{tab:v2-4gpu-5000}
\begin{tabular}{@{}l l l r r r r r r@{}}
\toprule
System & Task order & \shortstack[l]{Front\\camera} & Wall (s) & \shortstack{Replica\\loading (s)} & \shortstack{Eval.\\time (s)} & \shortstack{vs.\\official} & \shortstack{Steady\\steps/s} & \shortstack{Peak GPU\\mem.\ (GiB)} \\
\midrule
StarVLA official & one task per GPU & rendered & 63,068 & -- & 63,036 & 1.0$\times$ & -- & -- \\
\system{} & default & not rendered & 4,119 & 94 & 4,023 & 15.7$\times$ & 558 & 59.7--60.4 \\
\system{} & longest first & not rendered & 3,881 & 29 & 3,851 & 16.4$\times$ & 569 & 60.9--63.7 \\
\system{} & longest first & rendered & 4,295 & 32 & 4,262 & 14.8$\times$ & 511 & 61.8--62.6 \\
\system{}, $K60/g8$ & default & not rendered & 4,100 & -- & 3,985 & 15.8$\times$ & 575 & 68.9--69.3 \\
\bottomrule
\end{tabular}
\end{table}

\begin{table}[tbp]
\revcolor
\centering
\scriptsize
\caption{Successes per task over the 100 episodes of each RoboTwin Easy task, equal in the official evaluation and
all four \system{} launches, and the official evaluation's time for the task's 100 episodes on one GPU.}
\label{tab:v2-4gpu-5000-tasks}
\begin{tabular}{@{}l r r l r r@{}}
\toprule
Task & Succ. & Official (s) & Task & Succ. & Official (s) \\
\midrule
\code{adjust\_bottle} & 98 & 2,130 & \code{place\_can\_basket} & 66 & 6,061 \\
\code{beat\_block\_hammer} & 45 & 2,952 & \code{place\_cans\_plasticbox} & 88 & 4,984 \\
\code{blocks\_ranking\_rgb} & 50 & 8,164 & \code{place\_container\_plate} & 86 & 1,992 \\
\code{blocks\_ranking\_size} & 25 & 10,013 & \code{place\_dual\_shoes} & 32 & 6,630 \\
\code{click\_alarmclock} & 91 & 1,147 & \code{place\_empty\_cup} & 82 & 2,735 \\
\code{click\_bell} & 100 & 787 & \code{place\_fan} & 15 & 4,018 \\
\code{dump\_bin\_bigbin} & 86 & 3,267 & \code{place\_mouse\_pad} & 3 & 3,751 \\
\code{grab\_roller} & 93 & 1,495 & \code{place\_object\_basket} & 42 & 6,703 \\
\code{handover\_block} & 6 & 8,964 & \code{place\_object\_scale} & 4 & 3,968 \\
\code{handover\_mic} & 42 & 5,630 & \code{place\_object\_stand} & 47 & 2,977 \\
\code{hanging\_mug} & 1 & 11,488 & \code{place\_phone\_stand} & 0 & 4,095 \\
\code{lift\_pot} & 0 & 4,980 & \code{place\_shoe} & 53 & 3,550 \\
\code{move\_can\_pot} & 42 & 4,180 & \code{press\_stapler} & 81 & 1,674 \\
\code{move\_pillbottle\_pad} & 33 & 3,968 & \code{put\_bottles\_dustbin} & 20 & 17,828 \\
\code{move\_playingcard\_away} & 22 & 3,312 & \code{put\_object\_cabinet} & 49 & 6,729 \\
\code{move\_stapler\_pad} & 6 & 3,847 & \code{rotate\_qrcode} & 14 & 3,729 \\
\code{open\_laptop} & 50 & 5,027 & \code{scan\_object} & 24 & 5,337 \\
\code{open\_microwave} & 26 & 11,952 & \code{shake\_bottle} & 95 & 2,085 \\
\code{pick\_diverse\_bottles} & 27 & 4,325 & \code{shake\_bottle\_horizontally} & 95 & 2,076 \\
\code{pick\_dual\_bottles} & 42 & 3,972 & \code{stack\_blocks\_three} & 49 & 8,231 \\
\code{place\_a2b\_left} & 12 & 3,825 & \code{stack\_blocks\_two} & 80 & 4,042 \\
\code{place\_a2b\_right} & 14 & 3,923 & \code{stack\_bowls\_three} & 53 & 9,134 \\
\code{place\_bread\_basket} & 48 & 5,702 & \code{stack\_bowls\_two} & 94 & 3,988 \\
\code{place\_bread\_skillet} & 58 & 3,984 & \code{stamp\_seal} & 11 & 3,588 \\
\code{place\_burger\_fries} & 93 & 3,360 & \code{turn\_switch} & 20 & 3,926 \\
\midrule
\multicolumn{6}{@{}l}{Total: 2,313 successes; official 246,224\,s summed over episodes, 63,068\,s wall on four GPUs.} \\
\bottomrule
\end{tabular}
\end{table}
}

\subsection{Scalability}
\label{app:v2-scaling}
\paragraph{Scaling measurements.}
Each configuration in Table~\ref{tab:v2-h1-scaling} evaluates the full LIBERO-Long manifest once on the 20-core
host, with the same checkpoint and Policy as the cross-system comparison.

Table~\ref{tab:v2-h1-scaling} varies simulator workers $K$ and Policy replicas $g$ on one GPU.

\begin{table}[tbp]
\centering
\footnotesize
\setlength{\tabcolsep}{5pt}
\renewcommand{\arraystretch}{1.12}
\caption{LIBERO-Long single-GPU worker and Policy-replica scaling (one launch per configuration).}
\label{tab:v2-h1-scaling}
\begin{tabular}{@{}l r r r r r r r r@{}}
\toprule
Configuration & Successes & \shortstack{Eval.\\time (s)} & Steps/s & \shortstack{GPU\\util. (\%)} & \shortstack{Speedup\\vs. $K1/g1$} & \shortstack{Efficiency\\per worker} & \shortstack{Success\\agreement (\%)} & \shortstack{Terminal\\agreement (\%)} \\
\midrule
Search $\rightarrow$ $K12/g3$ & 447.7 & 679.0 & 202.5 & 97.0 & 5.3 & 0.44 & 97.1 & 78.5 \\
\midrule
$K1/g1$ & 451 & 3,559.6 & 38.4 & 15.4 & 1.0 & 1.00 & 100.0 & 100.0 \\
$K2/g1$ & 452 & 1,883.8 & 72.5 & 29.4 & 1.9 & 0.94 & 99.8 & 98.4 \\
$K4/g1$ & 451 & 1,213.6 & 112.5 & 45.7 & 2.9 & 0.73 & 98.4 & 87.0 \\
$K8/g1$ & 446 & 1,220.5 & 112.4 & 45.8 & 2.9 & 0.36 & 97.0 & 85.2 \\
$K12/g1$ & 446 & 1,216.1 & 113.7 & 46.3 & 2.9 & 0.24 & 97.8 & 78.6 \\
$K16/g1$ & 448 & 1,205.0 & 114.0 & 46.8 & 3.0 & 0.18 & 96.6 & 49.8 \\
$K20/g1$ & 447 & 1,221.3 & 112.0 & 45.6 & 2.9 & 0.15 & 96.0 & 42.4 \\
\midrule
$K4/g2$ & 453 & 1,001.8 & 136.1 & 59.7 & 3.6 & 0.89 & 98.8 & 88.0 \\
$K8/g2$ & 445 & 749.4 & 183.6 & 86.7 & 4.7 & 0.59 & 97.2 & 86.0 \\
$K12/g2$ & 449 & 748.0 & 183.3 & 87.7 & 4.8 & 0.40 & 97.2 & 77.4 \\
$K16/g2$ & 452 & 759.8 & 180.0 & 85.8 & 4.7 & 0.29 & 96.6 & 49.4 \\
\midrule
$K8/g3$ & 451 & 713.7 & 191.2 & -- & 5.0 & 0.62 & 98.8 & 86.8 \\
$K12/g3$ & 451 & 691.3 & 198.5 & -- & 5.1 & 0.43 & 97.6 & 78.4 \\
$K16/g3$ & 450 & 694.9 & 196.9 & -- & 5.1 & 0.32 & 95.8 & 48.6 \\
$K20/g3$ & 441 & 693.7 & 197.7 & -- & 5.1 & 0.26 & 94.4 & 41.8 \\
\midrule
$K12/g4$ & 448 & 676.9 & 203.8 & -- & 5.3 & 0.44 & 97.0 & 78.0 \\
\textbf{$K16/g4$} & 452 & \textbf{670.9} & 203.8 & -- & \textbf{5.3} & 0.33 & 96.6 & 48.6 \\
$K20/g4$ & 443 & 674.7 & 203.4 & -- & 5.3 & 0.26 & 95.2 & 41.8 \\
\bottomrule
\end{tabular}
\par\smallskip
\raggedright One RTX 4090 and a 20-core CPU quota on host~B. Same checkpoint, 500 episodes, and Policy as Table~\ref{tab:v2-c17-comparison}. The first row is autotuning with the diagonal search, three launches of \system{} (median search 506\,s, excluded from evaluation time): successes are the mean, time,
throughput and GPU utilization the median, speedup the geometric mean of the per-launch ratios, and
agreement the mean over launches. GPU utilization is the mean sampled utilization over the run. The $g=3$ and $g=4$ launches ran later on a second
instance of the same configuration, with the same runner and environment; their
speedup and agreement are against the same $K1/g1$ launch. Efficiency is speedup divided by $K$. Agreement pairs each run with $K1/g1$ on identical episodes: success outcome only, or success outcome and terminal step.\par
\end{table}

\section{Benchmark Compatibility}
\label{app:v2-compatibility}
\subsection{Supported Benchmark Configurations}
\begin{table}[!htb]
\centering
\small
\caption{Default work item and external assets of each adapter.}
\label{tab:v2-adapter-assets}
\renewcommand{\arraystretch}{1.15}
\begin{tabularx}{\linewidth}{@{}l >{\raggedright\arraybackslash}p{60mm} >{\raggedright\arraybackslash}X@{}}
\toprule
Benchmark & Default work item & External benchmark assets \\
\midrule
LIBERO & Episodes of one task & Pinned Python environment, simulator, dataset, initial states \\
LIBERO-Plus & Episodes of one variant & Pinned LIBERO-Plus environment, simulator archive, dataset, initial states \\
\rev{LIBERO-Pro} & \rev{One episode} & \rev{Pinned LIBERO-Pro environment and task files} \\
RoboTwin 2.0 & One task episode window & RoboTwin source and environment, checkpoint, validated seed cache \\
\rev{RoboDojo} & \rev{Up to 25 layouts of one task} & \rev{Isaac Sim environment, RoboDojo source and assets} \\
CALVIN & Consecutive sequences & CALVIN source, dataset, config, official 1,000-sequence manifest \\
RoboCasa & One task with multiple episodes & GR-1 simulator and task source, dataset, EGL environment \\
MetaWorld & One task with all its episodes & Pinned MT50 environment and protocol artifact \\
\rev{RLBench} & \rev{The whole suite, in PerAct's task order} & \rev{CoppeliaSim, RLBench source, PerAct's released test episodes} \\
\rev{robomimic} & \rev{One episode} & \rev{Pinned robosuite and robomimic environment} \\
\bottomrule
\end{tabularx}
\end{table}

Table~\ref{tab:v2-compatibility} in Section~\ref{sec:v2-implementation} lists the suites of each benchmark, and Table~\ref{tab:v2-adapter-assets} lists each adapter's default work item and the assets its benchmark supplies. The PyPI package supplies the Policy API, the Control Runtime, the Simulation Runtime, and the adapters; each benchmark supplies simulator
packages, datasets, manifests, render dependencies, and shared-storage assets.

\subsection{Failure Classification}
Run records distinguish contract mismatches, Policy failures, Simulation Runtime failures, resource
limits, record-integrity failures, and infrastructure errors, each tied to the stage that raised it.
Worker exits and allocator OOM errors map to requeue and replay; Policy-replica exits and
integrity errors abort the evaluation.

\end{document}